\documentclass{article} % For LaTeX2e
\usepackage[accepted]{icml2019arxiv}
\usepackage{booktabs}

\usepackage{amsmath,amsfonts,bm}

\def\eqref#1{equation~\ref{#1}}
\def\1{\bm{1}}

\DeclareMathAlphabet{\mathsfit}{\encodingdefault}{\sfdefault}{m}{sl}
\SetMathAlphabet{\mathsfit}{bold}{\encodingdefault}{\sfdefault}{bx}{n}

\newcommand{\Cov}{\mathrm{Cov}}
\DeclareMathOperator{\Tr}{Tr}

\usepackage{url}

\usepackage{breakurl}
\usepackage[breaklinks]{hyperref}

\usepackage{xcolor}       % blackboard math symbols
\usepackage{amsmath}
\usepackage{amsthm}
\usepackage{amssymb}
\usepackage{bbm}
\usepackage{wrapfig}
\usepackage{graphicx}
\usepackage[export]{adjustbox}
\usepackage{tablefootnote}

\makeatletter
\newtheorem*{rep@theorem}{\rep@title}
\newcommand{\newreptheorem}[2]{%
\newenvironment{rep#1}[1]{%
 \def\rep@title{#2 \ref{##1}}%
 \begin{rep@theorem}}%
 {\end{rep@theorem}}}
\makeatother

\newreptheorem{theorem}{Theorem}

\newreptheorem{proposition}{Proposition}

\icmltitlerunning{The Nonlinearity Coefficient}

\begin{document}

\twocolumn[
\icmltitle{The Nonlinearity Coefficient \\- Predicting Generalization in Deep Neural Networks}



\icmlsetsymbol{equal}{*}

\begin{icmlauthorlist}
\icmlauthor{George Philipp}{cmu}
\icmlauthor{Jaime G. Carbonell}{cmu}
\end{icmlauthorlist}

\icmlaffiliation{cmu}{School of Computer Science, Carnegie Mellon University, Pittsburgh, PA, USA}

\icmlcorrespondingauthor{George Philipp}{george.philipp@email.de}

\icmlkeywords{deep learning, neural networks, nonlinearity, activation functions, exploding gradients, vanishing gradients, architecture search, model selection}

\vskip 0.3in
]

\printAffiliationsAndNotice{}

\begin{abstract}
For a long time, designing neural architectures that exhibit high performance was considered a dark art that required expert hand-tuning. One of the few well-known guidelines for architecture design is the avoidance of exploding or vanishing gradients. However, even this guideline has remained relatively vague and circumstantial, because there exists no well-defined, gradient-based metric that can be computed {\it before} training begins and can reliably predict the performance of the network {\it after} training is complete.

We introduce what is, to the best of our knowledge, the first such metric: the nonlinearity coefficient (NLC). Via an extensive empirical study, we show that the NLC, computed in the network's randomly initialized state, is a powerful predictor of test error and that attaining a right-sized NLC is essential for attaining an optimal test error, at least in fully-connected feedforward networks. The NLC is also conceptually simple, cheap to compute, and is robust to a range of confounders and architectural design choices that comparable metrics are not necessarily robust to. Hence, we argue the NLC is an important tool for architecture search and design, as it can robustly predict poor training outcomes before training even begins.

\end{abstract}

\section{Introduction}

Designing neural architectures that perform well can be a difficult process. In particular, the exploding / vanishing gradient problem has been a major challenge for building very deep neural networks at least since the advent of gradient-based parameter learning \citep{hochreiterThesis,LSTM,RNNvanishingGradient}. However, there is still no consensus about which metric should be used for determining the presence of pathological exploding or vanishing gradients. Should we care about the length of the gradient vector \citep{heInit}, or about the size of individual components of the gradient vector \citep{depthScalesMeanField,resNetMeanField,normalizedInitialization}, or about the eigenvalues of the Jacobian \citep{orthogonalInitialization,explodingPascanu,eigenspectrum}? Depending on the metric used, different strategies arise for combating exploding and vanishing gradients. For example, manipulating the width of layers as suggested by e.g. \citet{resNetMeanFieldHacking,pyramidal} can greatly impact the size of gradient vector components but tends to leave the length of the entire gradient vector relatively unchanged. The popular He initialization for ReLU networks \citep{heInit} is designed to stabilize gradient vector length, whereas the popular Xavier initialization for tanh networks \citep{normalizedInitialization} is designed to stabilize the size of gradient vector components. While the papers cited above provide much evidence that gradient explosion / vanishing when defined according to some metrics is associated with poor performance when certain architectures are paired with certain optimization algorithms, it is often unclear how general those results are.

%Recently, \cite{expl} demystified many issues surrounding the exploding problem. They introduce the {\it gradient scale coefficient (GSC)}, the first metric for determining the presence of pathological exploding gradients that can be shown to directly lead to training difficulties. They also introduced the {\it collapsing domain problem}, where hidden layer activations corresponding to different datapoints become more and more similar with depth. Building on these ideas, 

We make the following core contributions.

\begin{enumerate}
\item We introduce the {\it nonlinearity coefficient (NLC)}, a gradient-based measurement of the degree of nonlinearity of a neural network (section \ref{derivationSection}).
\item We show that the NLC, computed in the networks randomly initialized state, is a powerful predictor of test error and that attaining a right-sized NLC is essential for achieving an optimal test error, at least in fully-connected feedforward networks (section \ref{nlcSection}).
\item We show that, by design, the NLC is not susceptible to a range of confounders that render many other metrics unreliable, such as changes to input scale, input bias and input dimensionality (section \ref{metricsSection}).
\end{enumerate}

We demonstrate the properties of the NLC via an extensive empirical study covering a wide range of network architectures (section \ref{nlcSection}). The scope of our experiments exceeds that of the vast majority of related work. We conduct 55.000 full training runs. As the NLC is also conceptually simple and cheap to compute, it is a useful guide for architecture design and search. Architectures with a sub-optimal NLC can be discarded a priori and computational resources don't have to be spent on training them. 

The NLC (defined in section \ref{derivationSection}) combines the covariance matrices of the network input and output with the input-output Jacobian into a scalar metric. Despite its simplicity, it is tied to many important properties of the network. It is a remarkably accurate predictor of the network's nonlinearity as measured by the relative diameter of the regions in input space that can be well-approximated by a linear function (section \ref{derivationSection} and figure \ref{sensi}). It is closely related to the nonlinearity of the individual activation functions used in the network and the degree to which they can be approximated by a linear function (section \ref{meaningSection}). It is tied to the susceptibility of the network's output to small random input perturbations.

\section{Notation and Terminology}

We define a neural network $f$ as a function of the input $x$. Both $x$ and the output $f(x)$ are vectors of fixed dimensionality $d_{\text{in}}$ and $d_{\text{out}}$ respectively. We assume a prediction framework, where the output is considered to be the prediction and the goal is to minimize the value of the `error' $e$ over this prediction and the label $y$, in expectation over some data distribution $\mathcal{D}$, i.e. we wish to minimize $\mathbb{E}_{(x,y)\sim \mathcal{D}}[e(f(x),y)]$. In this paper, $e$ is always classification error. During training, we replace $\mathcal{D}$ with the training set and $e$ with the surrogate loss function $\ell$, which in this paper is always softmax plus cross-entropy. Let $\mathcal{J}(x) := \frac{df(x)}{dx}$ or simply $\mathcal{J}$ be the Jacobian of the output with respect to the input $x$. Let $\bar{x} := \mathbb{E}_{x \sim \mathcal{D}} x$ be the expectation of the data input and $\bar{f} := \mathbb{E}_{x \sim \mathcal{D}} f(x)$ be the expectation of the network output induced by the data distribution. Similarly, let $\Cov_x := \mathbb{E}_{x \sim \mathcal{D}}[(x - \bar{x})(x - \bar{x})^T]$ denote the covariance matrix of the data input and $\Cov_f := \mathbb{E}_{x \sim \mathcal{D}}[(f(x) - \bar{f})(f(x) - \bar{f})^T]$ denote the covariance matrix of the output induced by the data distribution. 

\section{On the nonlinearity of neural networks - deriving the NLC} \label{derivationSection}

%\begin{wrapfigure}{r}{4.7cm}
%\includegraphics[width=4.7cm]{graphs/sinillu.pdf}
%\caption{bla}\label{osc}
%\end{wrapfigure} 

\paragraph{Defining nonlinearity} A key trait responsible for the success of neural networks is the fact that they are (generally) nonlinear. Linear models have low expressivity and thus underfit significantly on complex datasets such as \mbox{CIFAR10}. An important characteristic of linear models is that they are fully determined by the Jacobian and output value taken at a single input, i.e. for any fixed input $x'$, we have $f(x) = f(x') + [\mathcal{J}(x')](x - x')$. In other words, the local linear approximation induced by the Jacobian at any input is an exact representation of the model everywhere. This is not true for nonlinear models. The regions in input space where individual local linear approximations are accurate estimates of $f$ within some specified tolerance may be bounded. In fact, the more nonlinear a model becomes, the smaller these regions get.

We can easily formalize this if we assume that the input domain of $f$ is bounded and convex. Consider some input $x$ in such an input domain $D \subset \mathbb{R}^{d_\text{in}}$ and let $b_\text{in}$ be a point on the boundary of $D$. Then we can define the nonlinearity of $f$ at $x$ in the direction of $b_\text{in}$ to be the fraction of the distance from $x$ to $b_\text{in}$ in which the local linear approximation remains accurate. We can measure this accuracy in terms of whether the approximation remains within some tolerance of the function when projected onto a certain direction in output space. Formally, given some tolerance $T$ and vector $\delta_\text{out} \in \mathbb{R}^{d_\text{out}}$, we can say the nonlinearity of $f$ with respect to $(x,b_\text{in},\delta_\text{out},T)$ is the smallest $C \ge 1$ such that for all $c \le \frac{1}{C}$ we have $\frac{c}{T}\delta_\text{out}^T[\mathcal{J}(x)](b_\text{in} - x) \le \delta_\text{out}^T[f(x + c(b_\text{in} - x)) - f(x)] \le cT\delta_\text{out}^T[\mathcal{J}(x)](b_\text{in} - x)$.

In order to turn this into a practical definition for neural networks, we have to: (a) replace the bounded, convex domain with the data distribution $\mathcal{D}$ of arbitrary shape and (b) eliminate the need to manually specify $x$, $b_\text{in}$ and $\delta_\text{out}$. To do this, we will draw $x$ from $\mathcal{D}$ and draw $\delta_\text{out}$ from the unit Gaussian $\mathcal{N}(0,I_{d_\text{out}})$. Finally, we decide to model the random vector $b_\text{in} - x$ by drawing it from $\mathcal{N}(0, \Cov_x)$, i.e. a Gaussian that is fit to the data input distribution.

\paragraph{Definition 1.} We define the {\it nonlinearity distribution} of a network $f$ with respect to a data distribution $\mathcal{D}$ and tolerance $T$ as the distribution over $C$, where $C$ is the smallest value greater or equal to 1 such that for all $c \le \frac{1}{C}$ we have $\frac{c}{T}\delta_\text{out}^T[\mathcal{J}(x)]\delta_\text{in} \le \delta_\text{out}^T[f(x + c\delta_\text{in}) - f(x)] \le cT\delta_\text{out}^T[\mathcal{J}(x)]\delta_\text{in}$. The distribution over $C$ is induced by $x \sim \mathcal{D}$, $\delta_\text{in} \sim \mathcal{N}(0, \Cov_x)$,  and $\delta_\text{out} \sim \mathcal{N}(0, I_{d_\text{out}})$.

Linear functions achieve $C=1$ with probability 1. In general, $C \ge 1$. For the sake of clarity, we emphasize that this is not the ``correct'' definition of nonlinearity for neural networks or that other definitions are necessarily invalid. This definition is chosen to meaningfully capture the somewhat informal phenomenon of nonlinearity. It is used to derive and illustrate the NLC, and to show that the NLC captures a fundamental property of neural networks. %Note that we deliberately ignore the Hessian in our definition of nonlinearity as we do not think it is a useful metric for neural networks. See section TODO for details.

\paragraph{Defining the NLC} The nonlinearity distribution has limited practical utility. It is expensive to compute because each sample from the distribution requires a grid search over $c$. It is difficult to compute because of numerical stability issues induced by computing small differences. It is difficult to analyze because of its convoluted form. We now define the NLC as a scalar approximation to definition 1. The key insight behind the NLC and ultimately behind this paper is that there is a simple criterion for determining whether the local linear approximation can be accurate for a given $(x,\delta_\text{in},c)$: If $f(x) + c\mathcal{J}(x)\delta_\text{in}$ is far away from the codomain of $f(x)$, then because $f(x+c\delta_\text{in})$ lies in that codomain, $f(x) + c\mathcal{J}(x)\delta_\text{in}$ is far away from $f(x+c\delta_\text{in})$ and thus it is inaccurate. So an approximate lower bound for $C$ would be the ratio of the length of $\mathcal{J}(x)\delta_\text{in}$ and the distance from $f(x)$ to the boundary of the codomain. Specifically, we consider the quadratic mean of all lengths of $\mathcal{J}(x)\delta_\text{in}$, i.e. $\sqrt{\mathbb{E}_{x \sim \mathcal{D}, \delta_\text{in} \sim \mathcal{N}(0, \Cov_x)} ||\mathcal{J}(x)\delta_\text{in}||_2^2}$ and proxy the distance from $f(x)$ to the boundary of the codomain analogously by $\sqrt{\mathbb{E}_{\delta_\text{out} \in \mathcal{N}(0,\Cov_f)} ||\delta_\text{out}||_2^2}$. The ratio of both quantities equals the NLC as defined below.

\paragraph{Definition 2.} The {\it nonlinearity coefficient (NLC)} of a network $f$ and data distribution $\mathcal{D}$ is $NLC(f,\mathcal{D}) := \sqrt{\frac{\mathbb{E}_{x \sim \mathcal{D}}\Tr(\mathcal{J}(x)\Cov_x\mathcal{J}(x)^T)}{\Tr(\Cov_f)}}$.

As a sanity check, assume $f(x)$ is a linear function of form $Ax + b$. Then we have $\mathcal{J}(x) = A$ and $\Cov_f = \Cov(Ax + B) = A\Cov_xA^T$ and so $NLC(f, \mathcal{D}) = 1$ as desired. Note that an alternative interpretation of the NLC is that it represents the expected sensitivity of the network output with respect to small, randomly oriented changes to the input distributed according to a Gaussian fit to the data input distribution, normalized by the global variability of the network output. 

The NLC can be computed simply and cheaply. We describe the full algorithm in section \ref{computingSection} in the appendix. Finally, we refer readers interested in a pictorial illustration of the NLC to section \ref{pictorialSection}.

\section{On the predictive power of the NLC - large-scale empirical study} \label{nlcSection}

\paragraph{Architectures used} We sampled architectures at random by varying the depth of the network, the scale of initial trainable weights, scale of initial trainable bias vectors, activation function, normalization method, presence of skip connections, location of skip connections and strength of skip connections. We chose from a set of 8 activation functions (table \ref{nlinfo}), which were further modified by random dilation, lateral shift and debiasing. For now, we only considered fully-connected feedforward networks, as is (perhaps unfortunately) common in analytical studies of deep networks (e.g. \citet{orthogonalInitialization,gradientCorrelation,
depthScalesMeanField}). We have no reasons to suspect our results will not generalize to CNNs, and we plan to investigate this point in future work. See section \ref{architectureSection} for the full details of our architecture sampling scheme.

%\begin{figure}
%\includegraphics[width=\textwidth]{graphs/sensi.pdf}
%\caption{Relationship between the NLC and the relatize size of the region in which the gradient is informative in a random direction. The more red a point is, the higher $DBC(L)$. The bluer a point is, the stronger the skip connections.}\label{sensi}
%\end{figure}

\paragraph{Datasets used} We studied three datasets: MNIST, CIFAR10 and waveform-noise. All our results were highly consistent across these datasets. waveform-noise is from the UCI repository of datasets popular for evaluating fully-connected networks \citep{selu}. See section \ref{datasetsSection} for further details on dataset selection and preprocessing. We sampled 250 architectures per dataset, a total of 750.

\begin{wrapfigure}{r}{4.7cm}
\includegraphics[width=4.7cm]{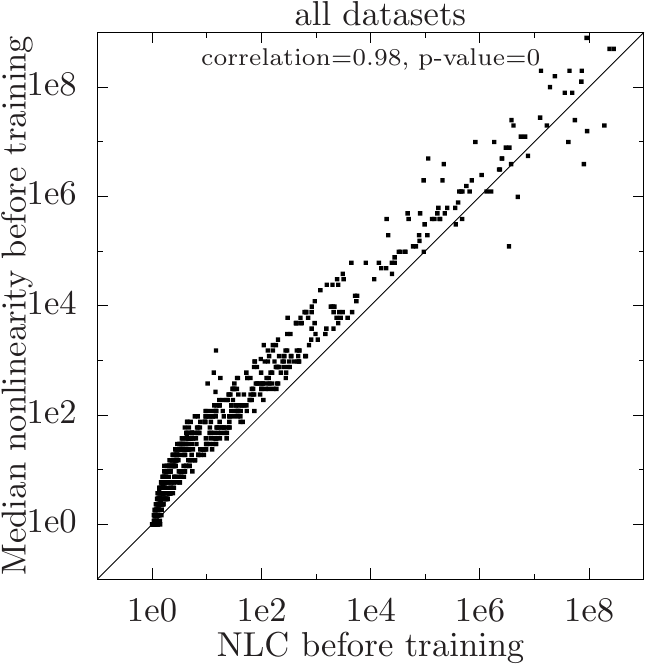}
\caption{NLC vs the median of the nonlinearity distribution. See section \ref{derivationSection} and \ref{sensiDetails} for details.}\label{sensi}
\end{wrapfigure} 

\paragraph{Training protocol} We trained each architecture with SGD with 40 different starting learning rates and selected the optimal one via a held-out validation set, independently for each architecture. During each run, we reduced the learning rate 10 times by a factor of 3. All training runs were conducted with 64 bit precision floating point computations. See section \ref{detailsSection} for further experimental details and section \ref{lratesection} for an analysis of how the learning rate search and numerical precision contributed to the outcome of our study.

\begin{figure*}
\includegraphics[width=0.95\textwidth]{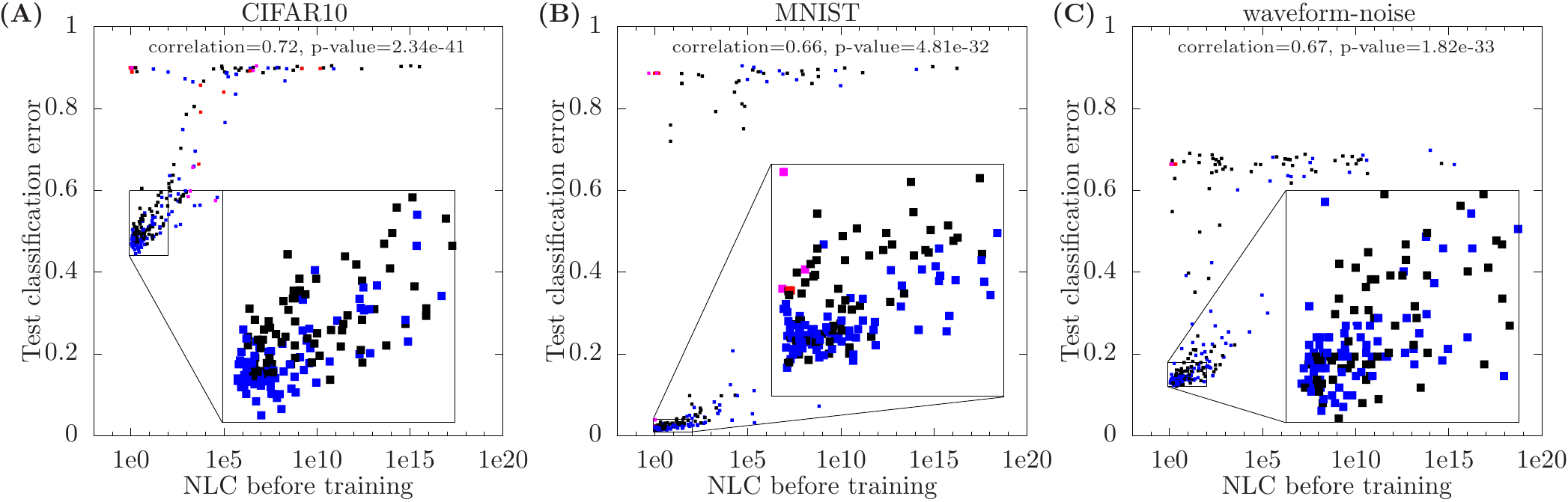}
\caption{NLC versus test error. Points shown in red correspond to architectures with high output bias ($\sqrt{\mathbb{E}_x||f||_2^2} > 1000\sqrt{\mathbb{E}_x||f-\bar{f}||_2^2}$). Points shown in blue correspond to architectures that have skip connections. Inset graphs in the bottom right are magnifications of the region $0.8 < NLC < 100$. See section \ref{nlcTestDetails} for details.} \label{NLCtest}
\end{figure*}

\begin{figure*}[!ht]
\centering
\includegraphics[width=0.74\textwidth]{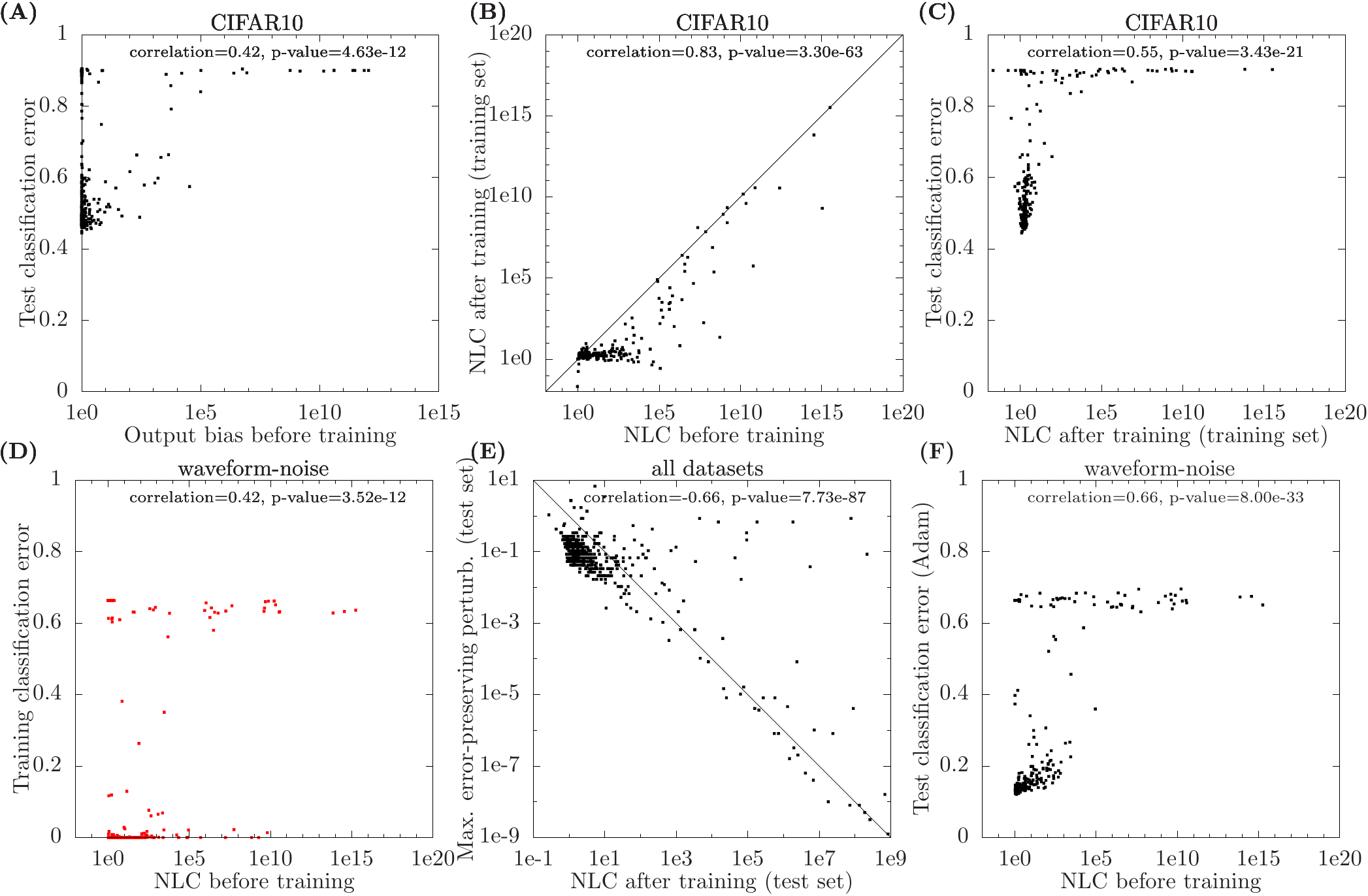}
\caption{Detailed results from our empirical study. See main text for explanation and sections \ref{nlcTestDetails} (figures A/B/C/D/F) and \ref{errorRobustnessDetails} (figure E) for further details.}\label{NLCcombo}
\end{figure*}

\paragraph{Presentation of results} The results of this study are shown in figures \ref{sensi}, \ref{NLCtest}, \ref{NLCcombo}, \ref{depth}, \ref{trainrun}, \ref{lrate}, \ref{bias} and \ref{skipsplit}. All figures except figure \ref{trainrun} are scatter plots where each point corresponds to a single neural architecture. In most graphs, we show the correlation and its statistical significance between the quantities on the x and y axis at the top. Note that if any quantity is plotted in log-scale, the correlation is also computed using the log of that quantity. For each architecture, we only studied a single random initialization. Given a limited computational budget, we preferred studying a larger number of architectures instead of multiple initializations. Note that all values depicted that were computed after training, such as test error or `NLC after training' in figure \ref{NLCcombo}C, are based on the training run which achieved the lowest validation classification error, as described above.

\paragraph{The NLC measures nonlinearity} First, we verify that the NLC is indeed an accurate measure of nonlinearity as defined by the nonlinearity distribution from definition 1. In figure \ref{sensi}, we plot the median of the nonlinearity distribution against the NLC in the randomly initialized state. We find a remarkably close match between both quantities. This shows empirically that our informal derivation of the NLC in the latter half of section \ref{derivationSection} leads to accurate nonlinearity estimates.

\paragraph{The NLC predicts test error} In figure \ref{NLCtest}, we plot the NLC computed in the randomly initialized state, {\it before} training, against the test error {\it after} training. We find that for all three datasets, the test error is highly related to the NLC. Further, figure \ref{NLCtest} indicates that one must start with an NLC in a narrow range, approximately between 1 and 3, to achieve an optimal test error, and the further one deviates from that range, the worse the achievable test error becomes. While we do not claim that this is the ideal range for all possible datasets, we find it to be consistent across the three datasets we studied. It is worth noting that some architectures, despite having an NLC in or close to this range, performed badly. One cause of this, high output bias, is explored later in this section. To verify that our results were not dependent on using the SGD optimizer, we re-trained all 250 waveform-noise architectures with Adam using the same training protocol. In figure \ref{NLCcombo}F, we find that the results closely match those of SGD from figure \ref{NLCtest}C.

\paragraph{NLC after training} In figure \ref{NLCcombo}B, we plot the value of the NLC before training versus after training. Both values were computed on the training set. We find that for the vast majority of architectures, the value of the NLC decreases. However, if the NLC is very large in the beginning, it remains so. Overall, the before-training NLC significantly predicts the after-training NLC. In figure \ref{NLCcombo}C, we plot the after-training NLC versus test error. We find that unless the NLC lies in a narrow range, test error is close to random. Interestingly, the after-training NLC has a significantly lower correlation with test error than the before-training NLC.

\paragraph{NLC predicts generalization, not necessarily trainability} In figure \ref{NLCcombo}D, we show the training error achieved by our architectures on waveform-noise. We re-trained all 250 architectures without using early stopping based on the validation error and considered an even larger range of starting learning rates. We depict the lowest training classification error that was achieved across all learning rates. Points are shown in red for visibility. We find that several architectures that have a very high NLC nonetheless achieve a zero or near-zero training error. This finding is somewhat contrary to that of \citet{depthScalesMeanField} and \citet{meanFieldCNN}, who report that networks which are highly sensitive to small input perturbations are untrainable. The reason we were able to train some of these architectures was our extensive learning rate search as well as our decision to train with 64 bit precision. In fact, we found that trainable architectures with high NLC generally require very small learning rates and very small parameter updates. One architecture required a learning rate as small as $4*10^{-19}$! See section \ref{lratesection} for further analysis on this point.

Of course, one might expect that a high sensitivity to small changes in the input leads to poor generalization. As a sanity check, we corrupted the test set with small random perturbations and measured how large these perturbations could be before the test error increased significantly. We plot this in figure \ref{NLCcombo}E. As expected, for the majority of high-NLC architectures, predictions can be corrupted and the error increased with incredibly small perturbations.

\paragraph{Summary} We interpret our results as follows. To generalize, the network must attain a critical NLC after training. This only occurs if the initial NLC is already close. In that case, the networks often learns automatically to adopt a more ideal NLC. However, unless the initial NLC is itself in the critical range, we cannot attain optimal performance.

%For space reasons, we only presented results for CIFAR10 in figures \ref{NLCcombo}A/B/C. Results from other datasets closely match those shown. See section \ref{furtherPlotsSection} for further plots, statistical tests, and discussion.

\paragraph{Further predictors: output bias and skip connections} In figure \ref{NLCtest}, we mark in red all points corresponding to architectures that have an output bias greater 1000, where we define output bias as $\sqrt{\frac{\mathbb{E}_x||f(x)||_2^2}{\mathbb{E}_x||f(x)-\bar{f}||_2^2}}$. All of these architectures attain a high test error, even if their NLC is small. In figure \ref{NLCcombo}A, we plot the output bias computed before training against test error. We find that indeed, to achieve an optimal test error, a low initial output bias is required. In section \ref{biasSection}, we further show that just as the NLC, the output bias also tends to decline during training and that attaining a very low output bias after training is essential. We also show that in contrast to a high NLC, a high output bias universally destroys trainability as well as generalization. We describe how to compute the output bias in section \ref{biasComputingSection}.

\begin{table*}
\centering
{
\scriptsize
\begin{tabular}{lcccccccc}
&ReLU&SELU&tanh&sigmoid&even tanh&Gaussian&square&odd square\\ \hline\hline
Formula&$\max(s,0)$&(see caption)&$\tanh(s)$&$\frac{1}{1+e^{-s}}$&$\tanh(|s|)$&$\frac{1}{\sqrt{2\pi}}e^{-\frac{s^2}{2}}$&$s^2$&$s*|s|$\\
Illustration&\includegraphics[scale=0.08,valign=c]{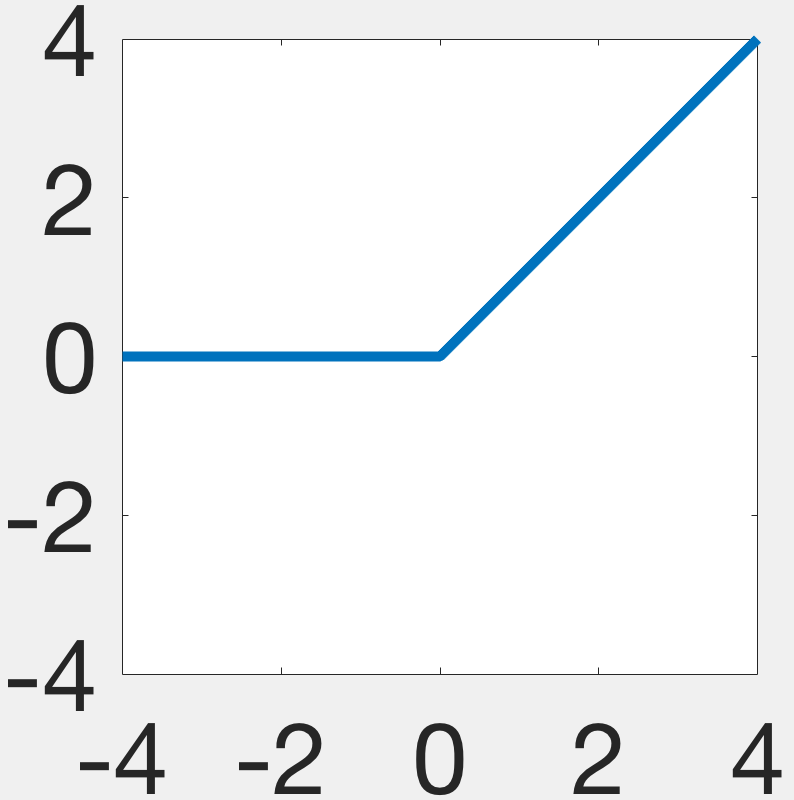}&\includegraphics[scale=0.08,valign=c]{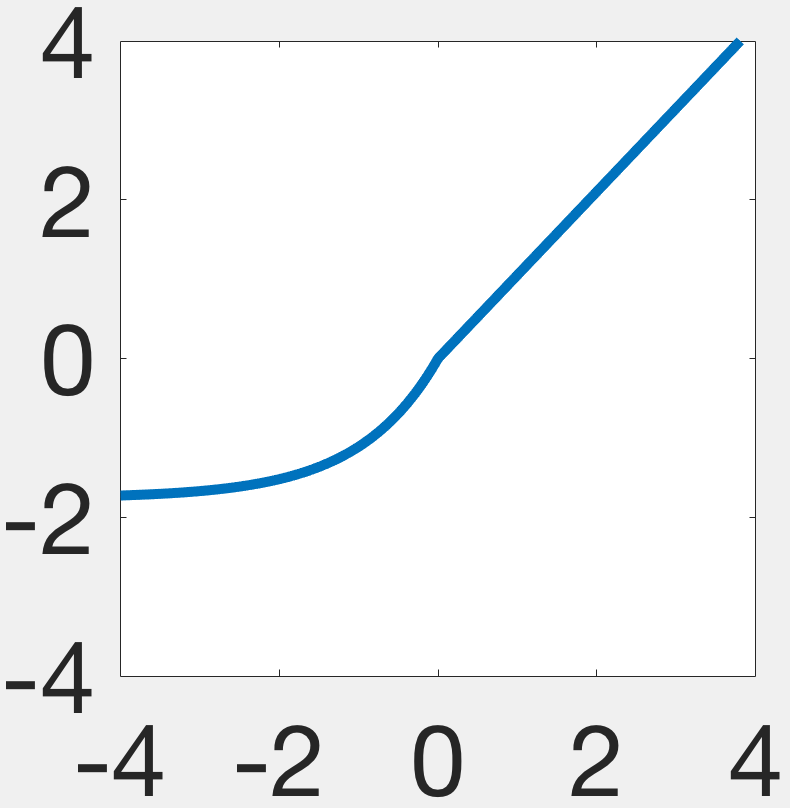}&\includegraphics[scale=0.08,valign=c]{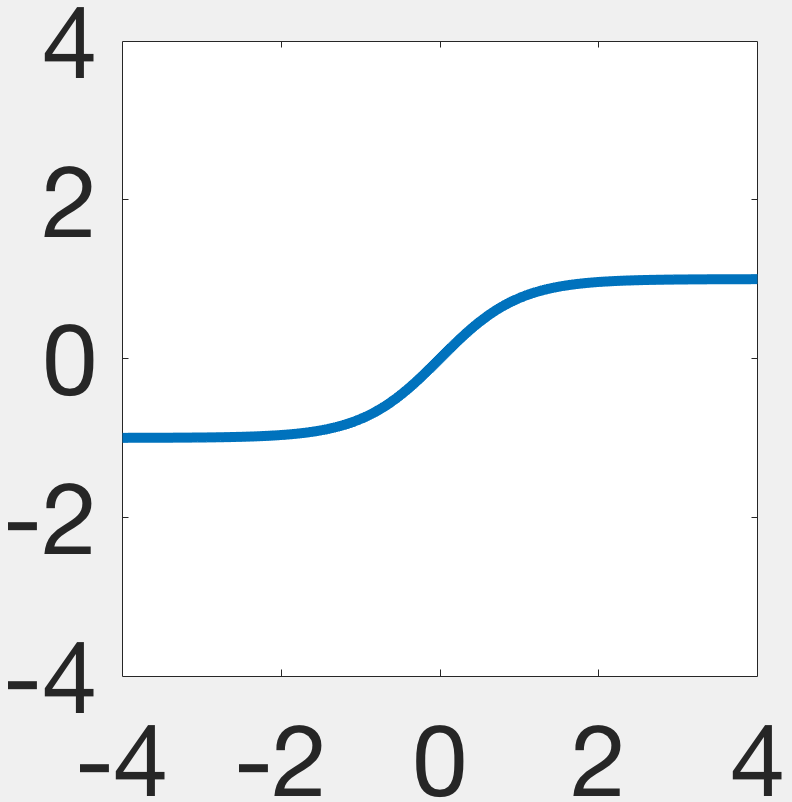}&\includegraphics[scale=0.08,valign=c]{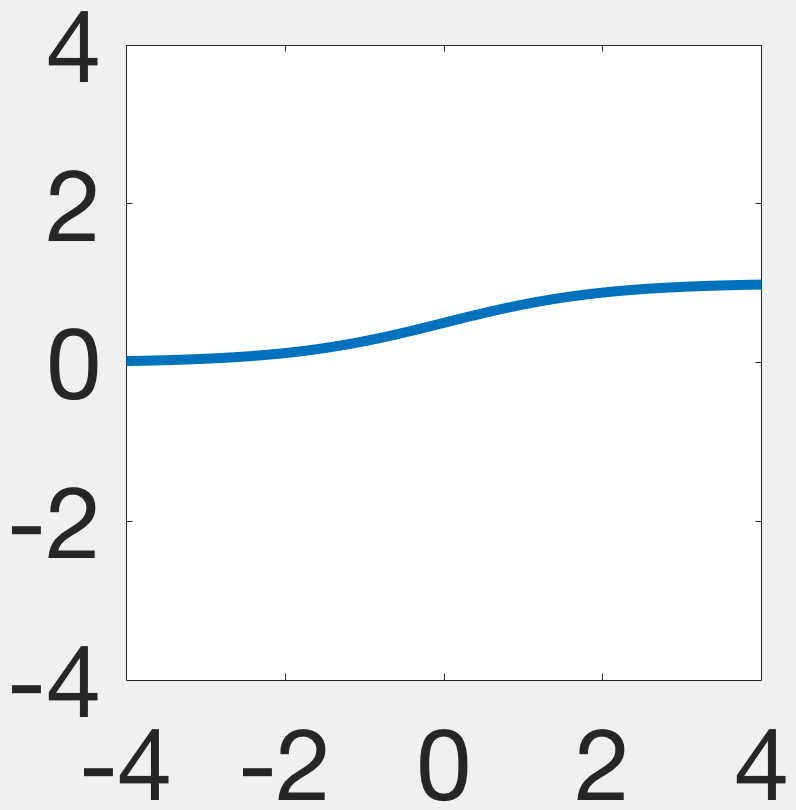}&\includegraphics[scale=0.08,valign=c]{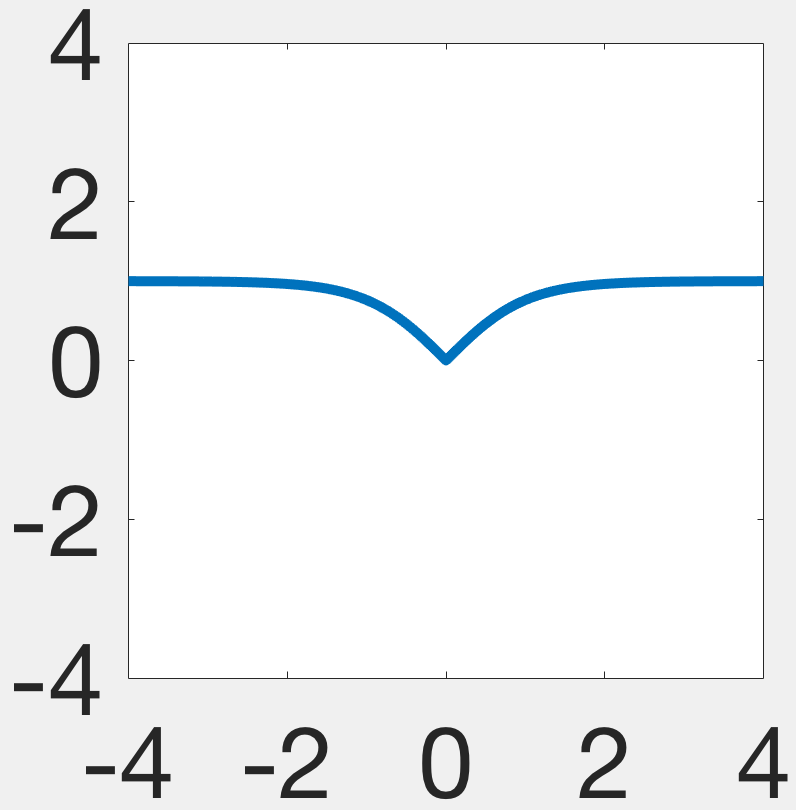}&\includegraphics[scale=0.08,valign=c]{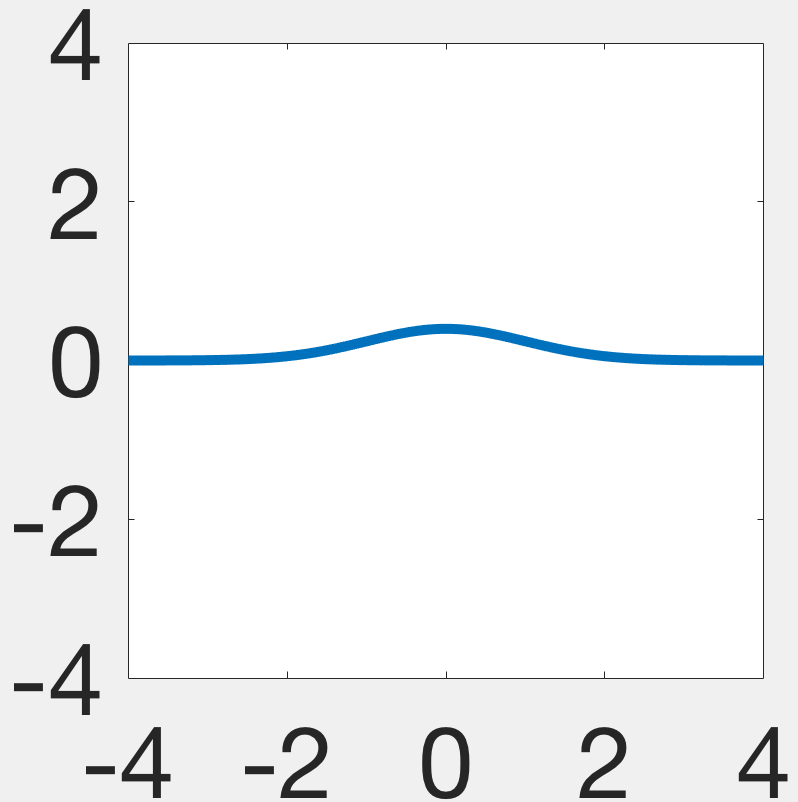}&\includegraphics[scale=0.08,valign=c]{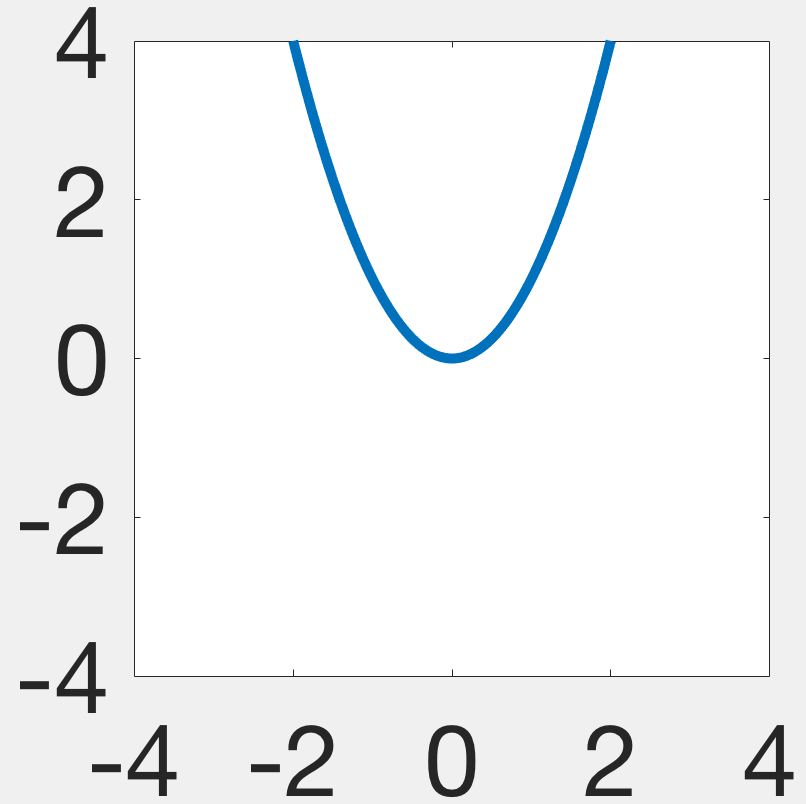}&\includegraphics[scale=0.08,valign=c]{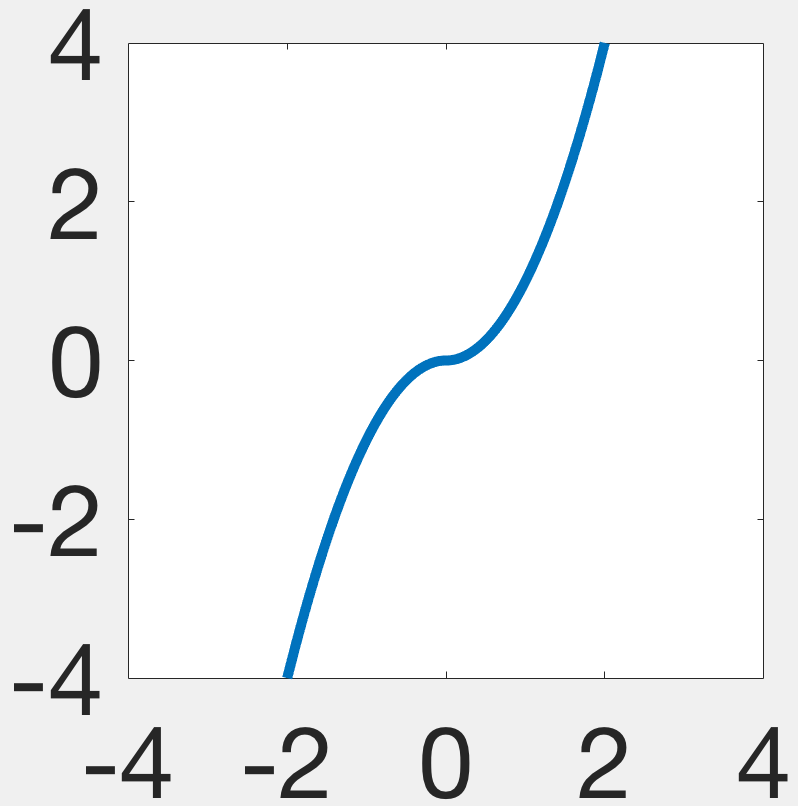}\\
(A) $NLC_\tau$&1.211&1.035&1.085&1.017&2.335&1.577&1.414&1.155\\
(B) NLC depth 2 (with batchnorm)&1.22&1.05&1.09&1.03&2.34&1.61&1.48&1.20\\
(C) $NLC_\tau^{48}$&$9793$&$5.21$&$50.19$&$2.25$&$4.76e17$&$3.13e9$&$1.67e7$&$1009$\\
(D) NLC depth 49 (with batchnorm)&$7376$&$7.06$&$59.9$&$3.06$&$4.90e17$&$3.74e9$&$4.28e12$&$1.38e11$\\
(E) Linear approximation error&0.222&0.030&0.075&0.0024&0.276&0.155&2.000&0.178\\
\end{tabular}
}
\caption{Activation functions used in this study with important metrics. See main text for explanation and section \ref{approximabilityDetails} for details. The formula for SELU is $\tau(s) = \mathbbm{1}_{s > 0}1.0507s + \mathbbm{1}_{s < 0}1.75814(e^s-1)$ \citep{selu}.}
\label{nlinfo}
\end{table*}

In figure \ref{NLCtest}, we plot in blue all points corresponding to architectures that have skip connections. \citet{expl} argued that skip connections reduce the gradient growth of general architectures as well as make a further contribution to performance. Correspondingly, we find that skip connections lead to lower NLC values and to lower test errors for a given NLC level. To enable a more convenient visual comparison, we plot results for architectures with and without skip connections separately in section \ref{skipsplitSection}.

%\begin{wraptable}{r}{4.5cm}
%\begin{tabular}{p{1.3cm}p{1.5cm}p{1.5cm}}
%NLC range&Least error&Rank\\\hline\hline
%$<\sqrt{10}$&&\\
%$[\sqrt{10},10]$&&\\
%$[10,10^{1.5}]$&&\\
%$[10^{1.5},10^{2}]$&&\\
%$[10^{2},10^{2.5}]$&&\\
%$[10^{2.5},10^{3}]$&&\\
%$[10^{3},10^{3.5}]$&&\\
%$[10^{3.5},10^{4}]$&&\\
%$[10^{4},10^{4.5}]$&&\\
%$[10^{4.5},10^{5}]$&&\\
%$>10^5$&&
%\end{tabular}
%\caption{Correlation and its statistical significance between NLC before training and test error, for each dataset.}\label{correlationTable}
%\end{wraptable}

%\begin{wrapfigure}{r}{4.7cm}
%\includegraphics[width=4.7cm]{graphs/NLCadam.pdf}
%\caption{NLC against test error. Networks were trained with Adam.}%\label{NLCadam}
%\end{wrapfigure} 

\section{On the linear approximability of activation functions} \label{meaningSection}

%\begin{wrapfigure}{r}{4.7cm}
%\includegraphics[width=4.7cm]{graphs/NLCsimple.png}
%\caption{The NLC measures the number of times a curve of the length of $f_l(s)$ can cross the domain of $f_l$, where $s$ is a line across the domain of $f_k$.}\label{nlcSimple}
%\end{wrapfigure} 

In this section, we expose the connection between the NLC and low-level properties of the activation functions the network uses. Given a 1d activation function $\tau$, we define $NLC_\tau := \sqrt{\frac{\mathbb{E}\tau'(s)^2}{(\mathbb{E}\tau(s)^2) -(\mathbb{E}\tau(s))^2}}$ where $s \sim \mathcal{N}(0,1)$. It is easy to check that if the input $x$ to a network is distributed according to a Gaussian with zero mean and identity covariance, and $f$ simply applies $\tau$ to each input component, then we have $NLC(f,\mathcal{D}) = NLC_\tau$. Consider a randomly initialized network where each layer is made up of a fully-connected linear operation, batch normalization, and an activation function $\tau$ that is applied component-wise. It turns out that if the network is sufficiently wide, the pre-activations of $\tau$ are approximately unit Gaussian distributed. This follows from the central limit theorem \citep{correlationLimit}. Hence, we expect each layer to contribute approximately $NLC_\tau$ to the NLC of the entire network. To verify this, we train a 2-layer network with batchnorm, which contains a single copy of $\tau$ at the single hidden layer. In table \ref{nlinfo}, we show $NLC_\tau$ for all 8 activation functions we used (line A), as well as the median empirical NLC over 10 random initializations of the 2-layer network (line B). We indeed find a close match between the two values. We then measure the NLC of 49-layer batchnorm networks, which contain 48 copies of $\tau$. For 6 out of 8 activation functions, this NLC (line D) closely matches the exponentiated $NLC_\tau^{48}$ (line C). Hence, we find that nonlinearity compounds exponentially and that the NLC of a network is closely tied to which activation function is used. Note that the reason that the NLC value of the `square' and `odd square' activation functions diverge from $NLC_\tau^{\text{depth}-1}$ at high depth is because those activation functions are unstable, which causes some inputs to grow in length with depth whereas the vast majority of inputs collapse towards the zero vector. 

%We explain this further in section \ref{furtherPlotsSection}.

We then verified that $NLC_\tau$ is a meaningful measure of nonlinearity for an activation function. We computed the best linear fit for each $\tau$ given unit Gaussian input and then measured the ratio of the power of the signal filtered by this best linear fit over the power of the preserved signal. In table \ref{nlinfo}(line E), we find that for ReLU, SELU, tanh, sigmoid and Gaussian activation functions, there is a close correspondence in that this linear approximation error is around $NLC_\tau - 1$. While this relationship breaks down for the 3 most nonlinear activation functions, their linear approximation error still exceeds those of the other 5. We conclude that $NLC_\tau$ is a meaningful measure of nonlinearity and that the NLC of an architecture can be calibrated by changing the linear approximability of the activation functions.

\section{On the robustness of the NLC vs other metrics - and related work} \label{metricsSection}

\begin{figure*}[!ht]
\centering
\includegraphics[width=\textwidth]{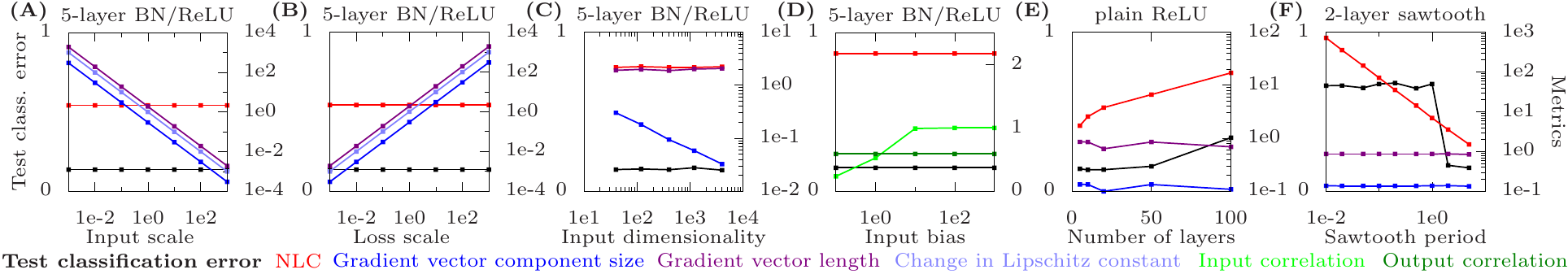}
\caption{The impact of confounders on test error (left y-axis) and various metrics (right y-axis). waveform-noise is the dataset. See section \ref{confDetails} for details.}\label{conf}
\end{figure*}

In this section, we discuss how the NLC compares against a range of metrics that are used for predicting performance or as guidelines for architecture design in the deep learning field. We find that each metric is susceptible to basic confounders that render them unreliable in practice. The NLC, by design, is not susceptible to these confounders.

%There is an enormous body of work on test error prediction for statistical models (e.g. \cite{statModelSelection1,statModelSelection3,
%statModelSelection4,statModelSelection5}). Discussing the applicability of classical statistical techniques to deep learning goes beyond the scope of this paper. As the NLC was developed specifically for use in deep learning, we will focus this section on the five metrics mentioned above.

\paragraph{Gradient vector component size}

Historically, there has not been a well-accepted metric for determining the presence of pathological exploding or vanishing gradients. Recently, \citet{depthScalesMeanField,resNetMeanField,normalizedInitialization} used the magnitude of gradient vector components for this job. We paraphrase this metric as $\sqrt{\mathbb{E}_{x}\frac{||\frac{d\ell}{dx}||_2^2}{d_\text{in}}}$ and abbreviate it as GVCS. It has several drawbacks that render it unreliable in practice.

%\begin{wrapfigure}{r}{4.7cm}
%\includegraphics[width=4.7cm]{graphs/osc.pdf}
%\caption{Comparing $GVCS(l)$ ({\color{red} red}), $NLC(l,L)$ (black) and the correlation of latent representations corresponding to uncorrelated inputs ({\color{blue} blue}) in a depth 10 batchnorm ReLU network with large weights, large biases and oscillating width. Note that we plot the observed values after each individual linear, batchnorm and ReLU operation in the order in which they appear.}\label{osc}
%\end{wrapfigure} 

The first drawback is that this metric is confounded by simple multiplicative rescaling. For example, assume we are using a network that begins with a linear operation followed by batch normalization or layer normalization \citep{layerNormalization}. Then we can re-scale the input data with an arbitrary constant $c$ and not only preserve the output of the network in the initialized state, but the entire trajectory of the parameter during training and therefore the final test error. Yet, multiplying the input by $c$ causes the GVCS to shrink by $c$. Thus we can arbitrarily control GVCS while preserving test error, and therefore GVCS is unreliable as a direct predictor of test error. In figure \ref{conf}A, we show this phenomenon for a 5 layer batchnorm-ReLU network. The x axis shows the constant with which the input was multiplied. This confounder can occur in practice as different users might scale their input data differently.

\begin{wrapfigure}{r}{4.7cm}
\includegraphics[width=4.7cm]{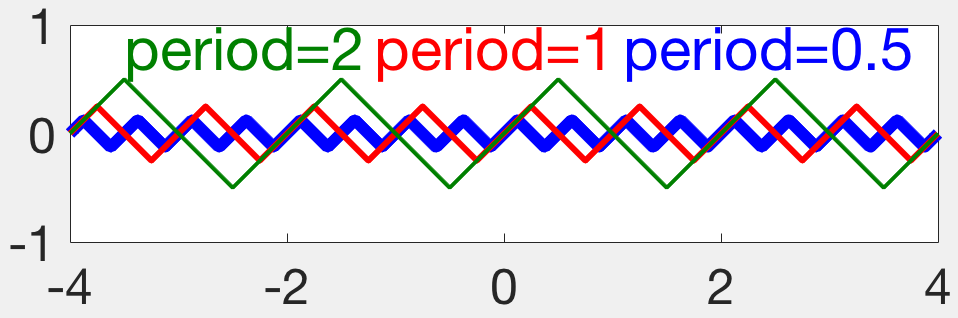}
\caption{Sawtooth activation function.}\label{saw}
\end{wrapfigure}

We observe a similar effect when the network loss is re-scaled with a constant $c$. This causes GVCS to grow by $c$. To preserve the learning trajectory, we only have to multiply the learning rate by $\frac{1}{c}$. We show this effect in figure \ref{conf}B with the same 5-layer network. This situation can occur when users choose loss functions which do not have a widely agreed-upon scale (e.g. some use $(x-y)^2$ while some use $\frac{1}{2}(x-y)^2$). It also occurs when users initialize weights differently, which can have an impact on the magnitude of the loss in e.g. plain ReLU networks where scaling factors are propagated forward.

The second drawback is that GVCS fails for highly nonlinear networks with stable gradients. In table \ref{conf}E, we show that as the depth of a plain He-initialized ReLU network increases, both the NLC and test error increase but GVCS remains stable. (The oscillation we see in figure \ref{conf}E is caused by the random initialization.) In fact, the He initialization was designed specifically to stabilize the gradient and is very popular in practice. However, we find that this is insufficient for ensuring high performance at high depth. At closer examination, we find that as the depth of plain ReLU networks increases, $\Tr(\Cov_f)$ decreases, which is captured by the NLC but not GVCS.

Similarly, consider the periodic sawtooth activation function shown in figure \ref{saw}. As the period length decreases, its $NLC_\tau$ converges to infinity as it becomes more and more erratic. Yet, GVCS is stable because $|\tau(s)'|=1$ holds independently of the period. Again, the nonlinearity is captured by the $\Tr(\Cov_f)$ term. We show results with 2-layer plain sawtooth networks of different periods in figure \ref{conf}F.

The third drawback is that the GVCS is also confounded by changing the input dimensionality. For example, consider a network that begins with a linear operation and has input dimensionality $d_\text{in}$. Then we can increase the input dimensionality by an integer factor $c$ by duplicating each input dimension $c$ times, which can happen in practice when e.g. considering images of different resolution. We can approximately maintain the learning dynamics by reducing the scale of initial weights of the first linear operator by $\sqrt{c}$ and the learning rate for that operator by $c$. Again, this transformation leaves the NLC unchanged but reduces GVCS by $\sqrt{c}$, allowing us to control GVCS while performance is unchanged once again. See figure \ref{conf}C for experimental results.

\paragraph{Gradient vector length / Lipschitz constant}

While less popular than GVCS, these two metrics are also used as an indicator of network performance (e.g. \citet{heInit} / \citet{parseval} respectively). Both metrics are susceptible to the same confounders as GVCS, except input dimensionality change. See figure \ref{conf}A/B/E/F for experimental results. Note that in figure \ref{conf}A/B we did not actually compute the Lipschitz constant but simply depict its relative change as the input scale / loss scale varies.

\paragraph{Input-output Jacobian}

Various measures of the Jacobian including Frobenius norm, mean eigenvalue and maximum eigenvalue have been considered as performance indicators (e.g. \citet{explosionGeneralization,eigenspectrum,explodingPascanu,orthogonalInitialization}).
While we will not discuss these measure in detail due to space limitations, they are closely related to GVCS / GVL and suffer from the same or highly similar confounders.

\paragraph{Correlation information}

Correlation information was recently proposed by \citet{depthScalesMeanField,resNetMeanField,meanFieldCNN,meanFieldRNN,meanFieldBN}. They claim that preserving the correlation of two inputs as they pass through the network is essential for trainability, and hence also for a low test error. We question the fundamental importance of preserving correlation as it can be easily confounded by the input bias. Assume we are using a network that employs batchnorm. Then biases in the features of the input do not significantly affect learning dynamics, as this bias will be removed by the first batchnorm operation. Yet, adding a constant vector to the input can arbitrarily increase correlation between inputs without affecting the correlation of the outputs. So, again, the degree of correlation change through the network can be manipulated arbitrarily without altering network performance. In figure \ref{conf}D, we show results for 5-layer ReLU batchnorm networks where an input bias was introduced by adding a fixed scalar constant $c$ to each component of the input. Such biases can occur in practice depending on how users normalize their data. The correlations in figure \ref{conf}D are the quadratic mean of pairwise correlations of inputs and outputs respectively.

\paragraph{Correlation depth scale} \citet{depthScalesMeanField,resNetMeanField,meanFieldCNN,meanFieldRNN,meanFieldBN} propose `correlation depth scale' as a specific metric for determining the preservation of correlation information. This metric measures the convergence rate of the correlation of two inputs to its limit in a hypothetical, infinitely-deep network. The authors claim that networks for which this convergence is sub-exponential are ideal, and they call those networks ``on the edge of chaos''. However, both the plain ReLU network from figure \ref{conf}E and the plain sawtooth network from figure \ref{conf}F lie at the edge of chaos. The former fails at high depth and the latter even fails at depth 2 when the sawtooth period is short. In both cases, while the convergence rate is sub-exponential in the limit, it is fast to very fast in the short term. This is not captured by the depth scale metric. On the other hand, the NLC is a metric of real-world, finitely-deep networks rather than hypothetical, infintiely-deep ones.

%As with the GVCS, the correlation of latent representations can vary wildly from operation to operation. In figure \ref{osc}, we plot the correlation of two uncorrelated inputs as they pass through the network. As the inputfs pass through the biased linear operation, correlation spikes, but this is spurious.

\begin{wrapfigure}{r}{4.7cm}
\includegraphics[width=4.7cm]{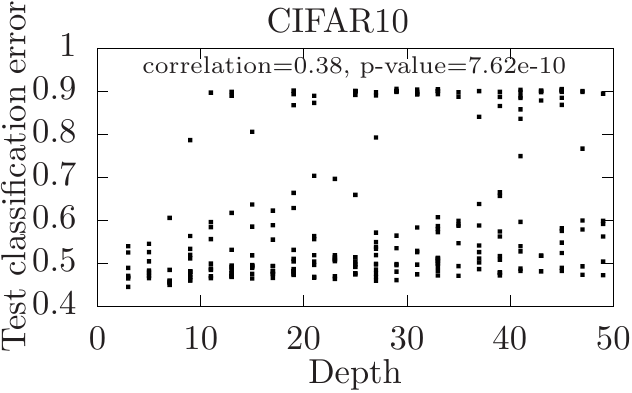}
\caption{Depth versus test error.}\label{depth}
\end{wrapfigure}

\paragraph{Depth} A large body of work has detailed the benefits of depth in neural networks (e.g. \citet{depth1,depth2,depth3,depth4,depth5,depth6,depth7}). Most of these works focus on finding specific functions which can be represented easily by deep networks, but require a prohibitively large number of neurons to represent for a shallow network. In figure \ref{depth}, we plot the test error achieved by our architectures on CIFAR10 against depth. We find that there is actually a positive correlation between both quantities, i.e. deeper networks tend to perform worse. We suspect this is mainly because deeper networks tend to have a larger NLC. 

%We form a bold hypothesis: When fully-connected networks are applied to practical learning problems, network depth has little to no impact on performance in general beyond its indirect impact via the NLC and the total number of parameters. While we do not have enough evidence for this conclusion, we believe this is an interesting point of investigation.

\paragraph{Other papers} Finally, we want to point to two papers that conducted large-scale studies similar to ours where test error was predicted across a variety of networks \citep{explosionGeneralization,marginPrediction}. We signficantly improve upon those papers because we compute the NLC {\it before} training begins whereas previous work predicts test error from metrics computed after training is complete. Not only is there much greater practical utility in pre-training prediction, but it also much more difficult as we do not have access to the network weights that produce the actual error.

\section{Discussion and conclusion}

We introduced the nonlinearity coefficient, a measure of neural network nonlinearity that is closely tied to the relative diameter of linearly approximable regions in the input space of the network, to the sensitivity of the network output with respect to small input changes, as well as to the linear approximability of activation functions used in the network. Because of this conceptual grounding, because its value in the randomly initialized state is highly predictive of test error while also remaining somewhat stable throughout training, because it is robust to simple network changes that confound other metrics such as raw gradient size or correlation information, because it is cheap to compute and conceptually simple, we argue that the NLC is the best standalone metric for predicting test error in fully-connected feedforward networks. It has clear applications to neural architecture search and design as it allows sub-optimal architectures to be discarded before training. In addition to a right-sized NLC, we also found that avoiding excessive output bias and using skip connections play important independent roles in performance.

This paper makes important contributions to several long-standing debates. We clearly show that neural networks are capable of overfitting when the model is too complex. In fact, our random architecture sampling scheme shows that such architectures are not rare. However, overfitting appears to be tied not to depth or the number of parameters, but rather to nonlinearity. In contrast to \citet{depthScalesMeanField,meanFieldCNN}, we find that while a very high output sensitivity harms generalization, it does not necessarily harm trainability. This difference is likely caused by our very extensive learning rate search and 64 bit precision training.

While the popular guidance for architecture designers is to avoid exploding and vanishing gradients as measured by e.g. GVCS, or more recently to choose networks ``on the edge of chaos'', we argue that achieving an ideal NLC is the more succinct criterion. This is not to say that any of the metrics discussed in section \ref{metricsSection} can't be highly predictive of test error in specific situations. In fact, the majority of these metrics would have significant predictive value within the empirical study conducted in this very paper. It is the solid conceptual grounding of the NLC that sets it apart. The NLC is not a superficial quantity, but is linked to a deep and robust property of neural networks. Its robustness to confounders is not merely an incidental advantage, but a testament to its careful design.

It has been argued that the strength of deep networks lies in their exponential expressivity (e.g. \citet{trajectoryTransitions,depth6}). While we show that the NLC indeed exhibits exponential behavior, we find this property to be largely {\it harmful}, not helpful, as did e.g. \citet{depthScalesMeanField}. While very large datasets may benefit from more expressivity, in our study such expressivity only leads to lack of generalization rather than greater trainability. In fact, at least in fully-connected networks, we conjecture that great depth doesn't confer significant practical benefit.

In future work, we plan to study whether the ideal range of NLC values we discovered for our three datasets ($1 \lessapprox NLC \lessapprox 3$) holds also for larger datasets and if not, how we might predict this ideal range a priori. We plan to investigate additional causes for why certain architectures perform badly despite a right-sized NLC, as well as extend our study to convolutional and densely-connected networks. We are interested in studying the connection of the NLC to e.g. adversarial robustness, quantizability, sample complexity, training time and training noise. Finally, unfortunately, we found the empirical measurement of the NLC to be too noisy to conclusively detect an underfitting regime. We plan to study this regime in future.

\bibliography{citations}
\bibliographystyle{icml2019}

\appendix
\newpage

\section{The NLC explained pictorially} \label{pictorialSection}

\begin{wrapfigure}{r}{4.7cm}
\includegraphics[width=4.7cm]{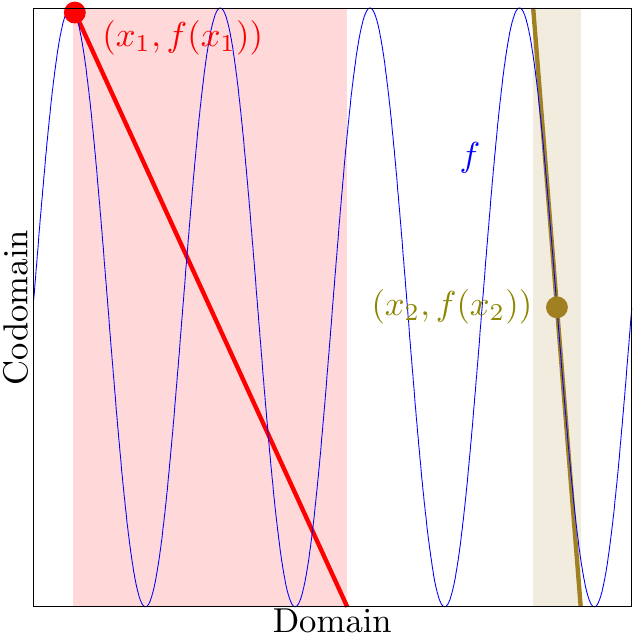}
\caption{1d pictorial illustration of the NLC}\label{sinillu}
\end{wrapfigure} 

The goal of this section is to provide an intuitive, graphical explanation of the NLC in addition to the mathematical derivation and analysis in section \ref{derivationSection} for readers interested in developing a better intuition of this concept.

\subsection{1D toy example}

In figure \ref{sinillu}, we illustrate the meaning of the NLC in the case of an example function $f$ with a single input and output dimension, and a bounded domain $\mathbb{D}$ and codomain $\mathbb{F}$. $f$ is a simple sin curve, shown in blue. $x_1$ and $x_2$ are two sample inputs. We plot the location of $(x_1,f(x_1))$ in red and the location of $(x_2,f(x_2))$ in olive. The thick red and olive lines correspond to the local linear approximation of $f$ at $x_1$ and $x_2$ respectively, which is simply the tangent line of the blue curve. The shaded olive and red regions correspond to the intervals in which the local linear approximations fall inside the codomain $\mathbb{F}$.

It is easy to check that the proportion of the domain covered by the red interval and olive interval is $\frac{\text{diameter}(\mathbb{F})}{f'(x_1)\text{diameter}(\mathbb{D})}$ and $\frac{\text{diameter}(\mathbb{F})}{f'(x_2)\text{diameter}(\mathbb{D})}$ respectively. The insight behind the NLC is that both linear approximations can only be accurate while they remain inside their respective shaded area, or at least close to it. This is evidently true in both cases, as both tangent lines quickly move away from the codomain outside the shaded region. In the case of $x_2$, this bound is also tight as the tangent tracks $f$ closely everywhere in the olive region. However, in the case of $x_1$, the bound is loose, as the red line completely decouples from $f$ throughout a large part of the red region. 

We can view $\frac{\text{diameter}(\mathbb{F})}{f'(x)\text{diameter}(\mathbb{D})}$ as the fraction of the domain covered by a single shaded region. The inverse value, $\frac{f'(x)\text{diameter}(\mathbb{D})}{\text{diameter}(\mathbb{F})}$, can be viewed as the number of shaded regions required to cover the entire domain. The NLC is simply the generalization of this concept to multiple dimensions, where the diameter is proxied by the covariance matrix, the gradient becomes the Jacobian, and the expectation is taken over the data distribution. It is worth noting that the NLC attempts to measure the ratio of diameter of domain and linearly approximable region, {\it not} the ratio of volumes. Informally speaking, the number of linearly approximable regions required to cover the domain behaves as $NLC^{d_\text{in}}$.

\subsection{2D neural network example}

\begin{table*}
\centering
{
\begin{tabular}{lccccccc}
depth & 2 & 5 & 10 & 15 & 20 & 25 & 50\\
 \hline\hline
NLC&1.22&2.25&5.97&15.2&37.7&95.8&9952\\
Illustration&\includegraphics[scale=0.15,valign=c]{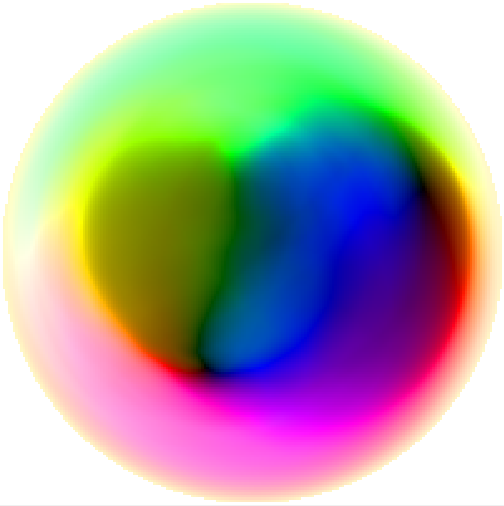}&\includegraphics[scale=0.095,valign=c]{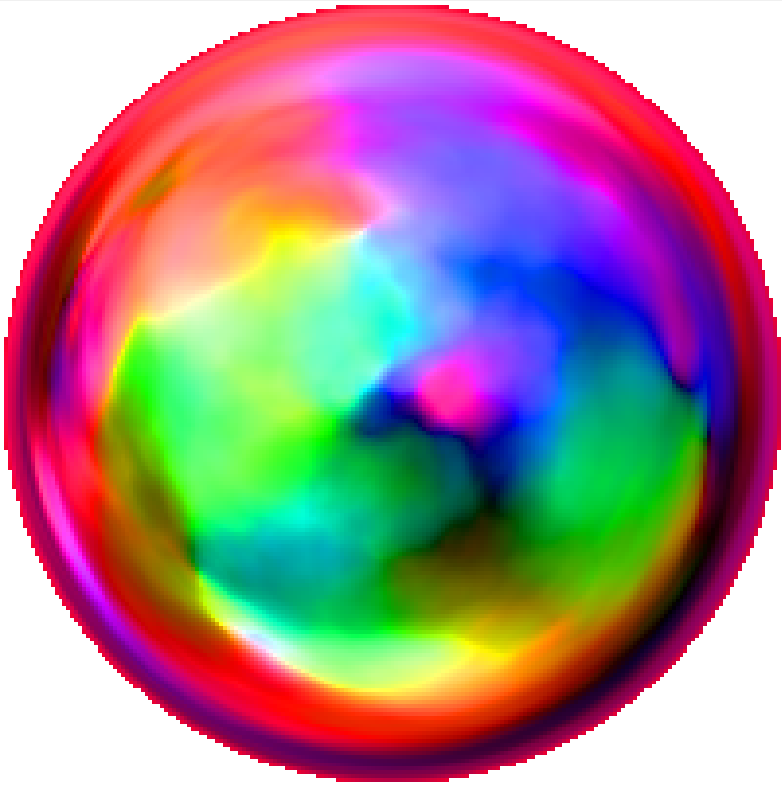}&\includegraphics[scale=0.14,valign=c]{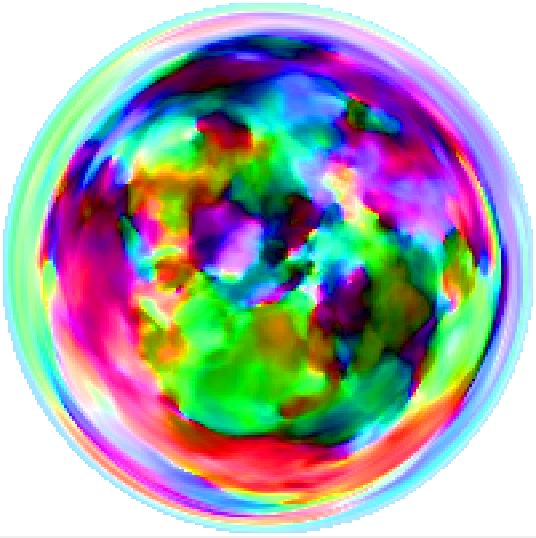}&\includegraphics[scale=0.11,valign=c]{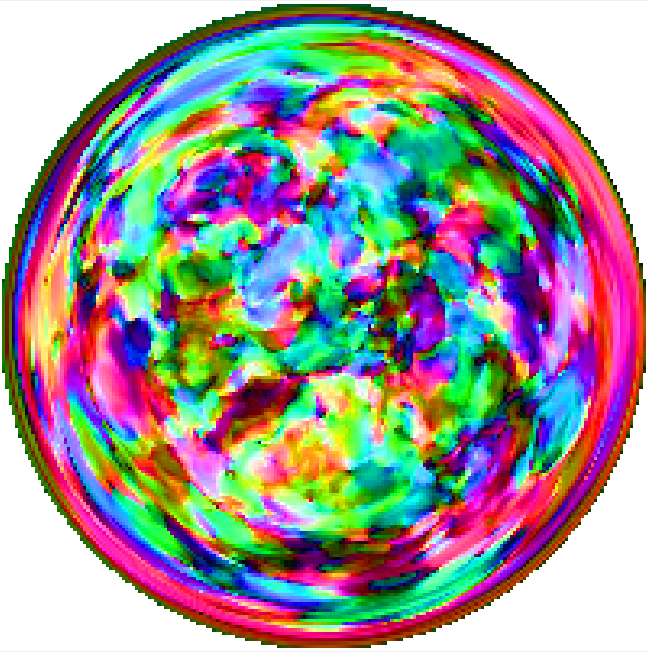}&\includegraphics[scale=0.090,valign=c]{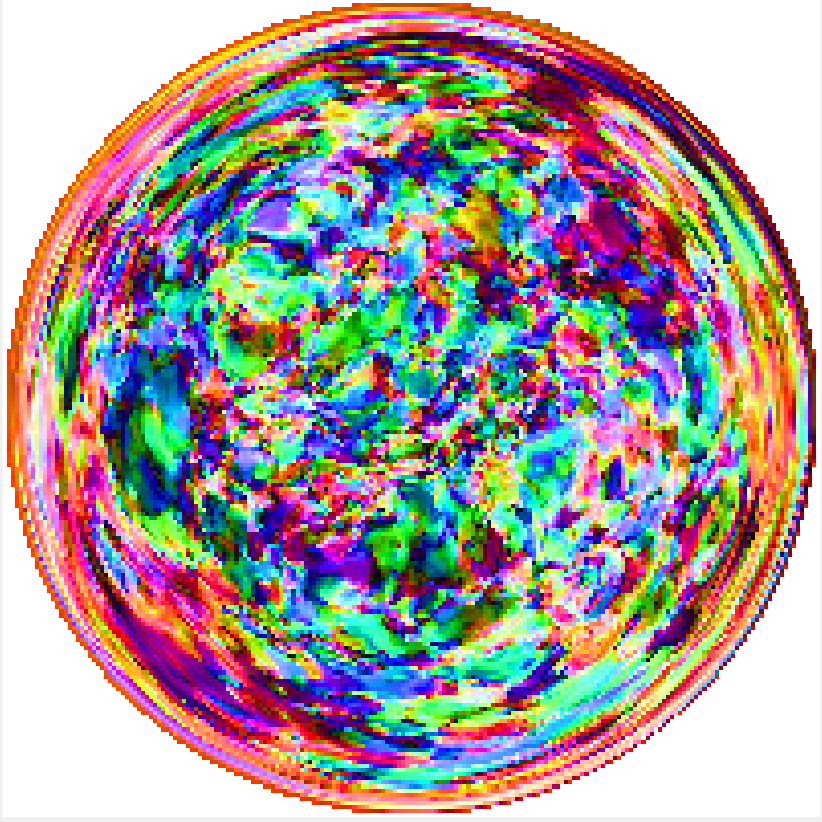}&\includegraphics[scale=0.10,valign=c]{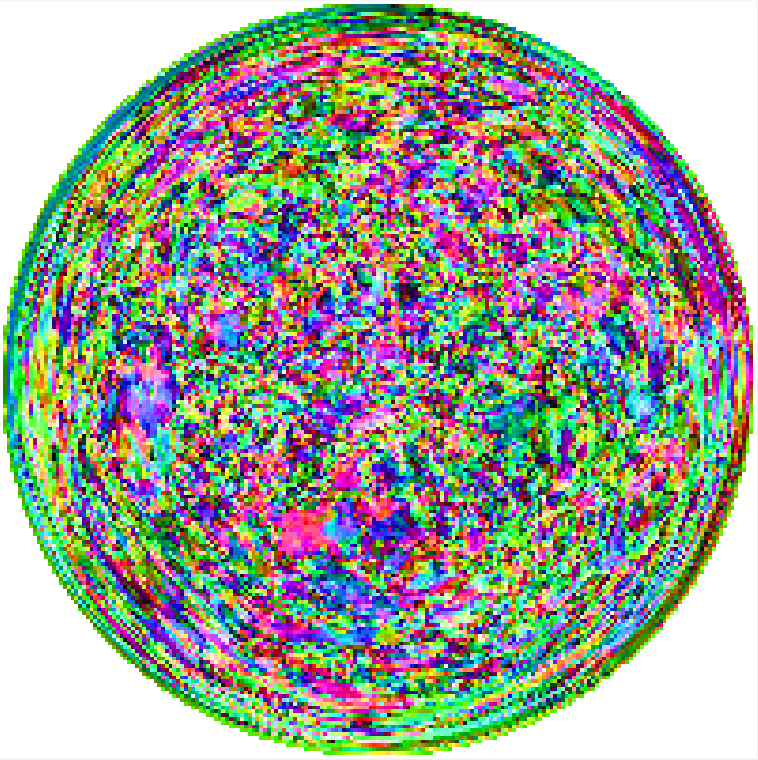}&\includegraphics[scale=0.14,valign=c]{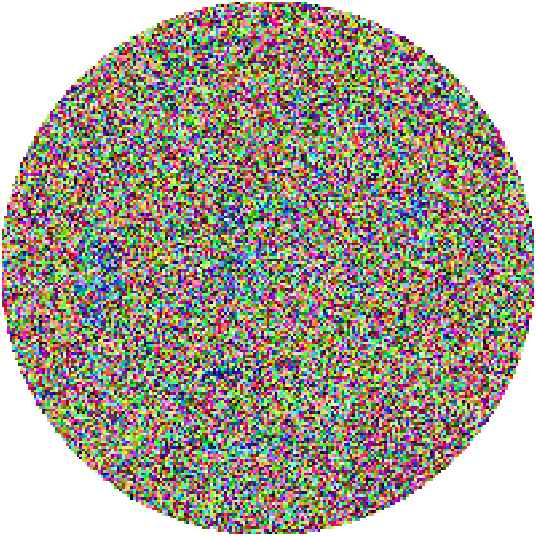}
\end{tabular}
}
\caption{Illustration of the function computed by fully-connected batchnorm-ReLU networks at different depths in the randomly initialized state. Each disc represents a 2D subspace of the input space and each color corresponds to a different region of the output space. CIFAR10 was used to compute the NLC.}
\label{illu}
\end{table*}

In this section, we illustrate the function computed by neural networks at varying levels of nonlinearity. Specifically, in Table \ref{illu}, we depict the function computed by fully-connected, He-initialized batchnorm-ReLU networks at seven different depths in their randomly initialized state. 

We set $d_\text{out}=3$ and set the width of all other layers to 100. We then generated three 100-dimensional random inputs $x^{(1)}$, $x^{(2)}$ and $x^{(3)}$ drawn from $\mathcal{N}(0,I_{100})$. We associated each point $(a,b,c)$ that lies on the unit sphere in $\mathbb{R}^3$, i.e. that has $a^2+b^2+c^2=1$, with the input $ax^{(1)} + bx^{(2)} + cx^{(3)}$. We call the sphere of points $(a,b,c)$ associated with these inputs the ``input sphere''.

\begin{wrapfigure}{r}{4.7cm}
\includegraphics[width=4.7cm]{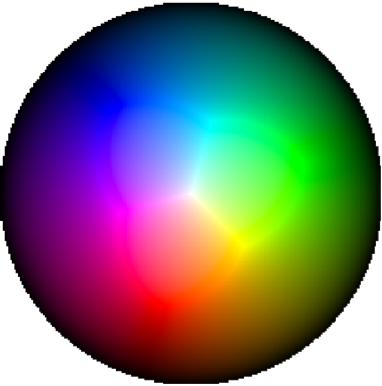}
\caption{Coloring of the output sphere used for the illustrations in table \ref{illu}, shown as an azimuthal projection.}\label{outputSphere}
\end{wrapfigure} 

We propagate each of those inputs forward through the network. We obtain a 3-dimensional output, which we divide by its length. Now the output lies on the unit sphere in $\mathbb{R}^3$. Each point on that ``output sphere'' is associated with a color as shown in figure \ref{outputSphere}. Finally, we color each point on the input sphere according to its respective color on the output sphere. 

These colored input spheres are shown in table \ref{illu} as azimuthal projections. The RGB values of colors on the output sphere are chosen so that the R component is largest whenever the first output neuron is largest, the G component is largest whenever the second output neuron is largest and the B component is largest whenever the third output neuron is largest. If we imagine that the output is fed into a softmax operation for 3-class classification, then ``purer'' colors correspond to more confident predictions.

For comparison, we show the NLC on CIFAR10 for batchnorm-ReLU networks of the same depth (median of 10 random initializations). We find that as depth and the NLC of the network increases, the color, and thus the value of the output, change more quickly as we move across the input space. This chaotic behavior of the output correspondingly implies smaller linearly approximable regions.

\section{Large-scale empirical study - additional results}

In this section, we expand upon findings from our large-scale empirical study that were outlined in section \ref{nlcSection}.

\subsection{What is the ideal learning rate?} \label{lratesection}

\begin{figure*}
\includegraphics[width=\textwidth]{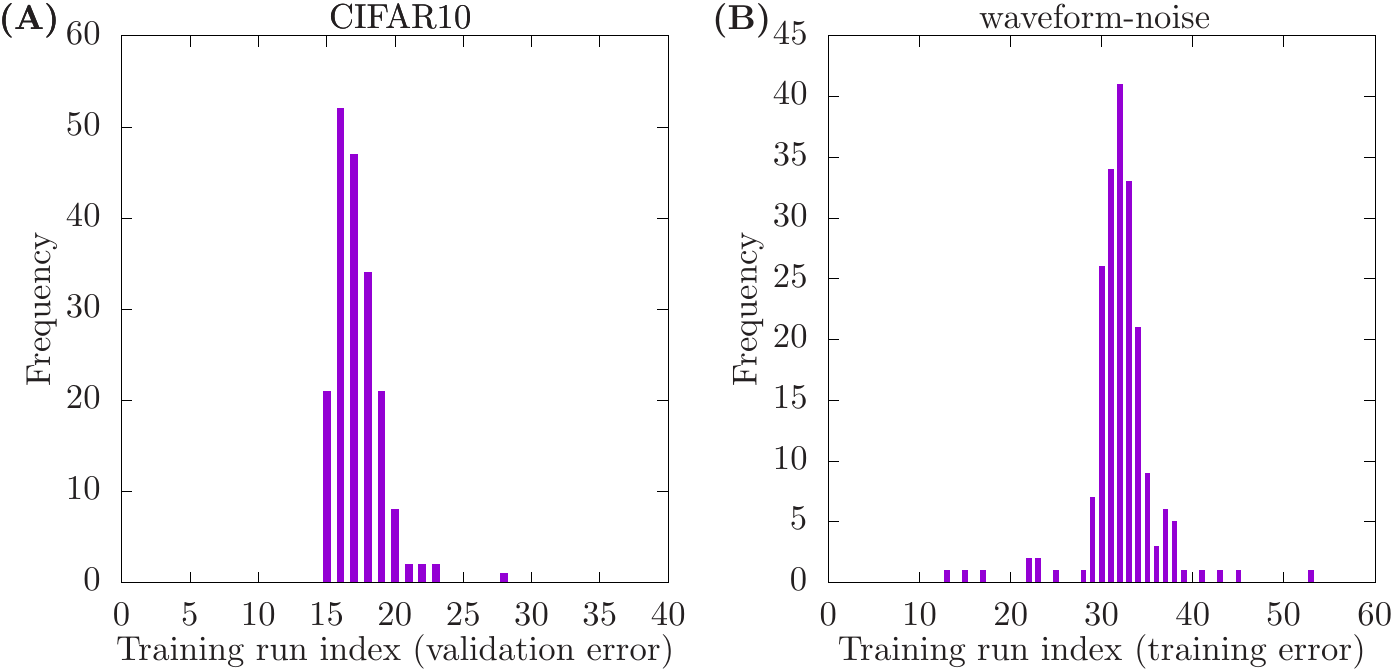}
\caption{Frequency with which each training run minimized the validation error on CIRAR10 (A) / training error on waveform-noise (B). Note: Architectures which did not achieve a better-than-random validation error were omitted in (A) and architectures that did not achieve a better-than-random training error were omitted in (B). We set those thresholds at 80\% for CIFAR10 (10 different labels) and 50\% for waveform-noise (3 different labels).}\label{trainrun}
\end{figure*}

\begin{figure*}
\includegraphics[width=\textwidth]{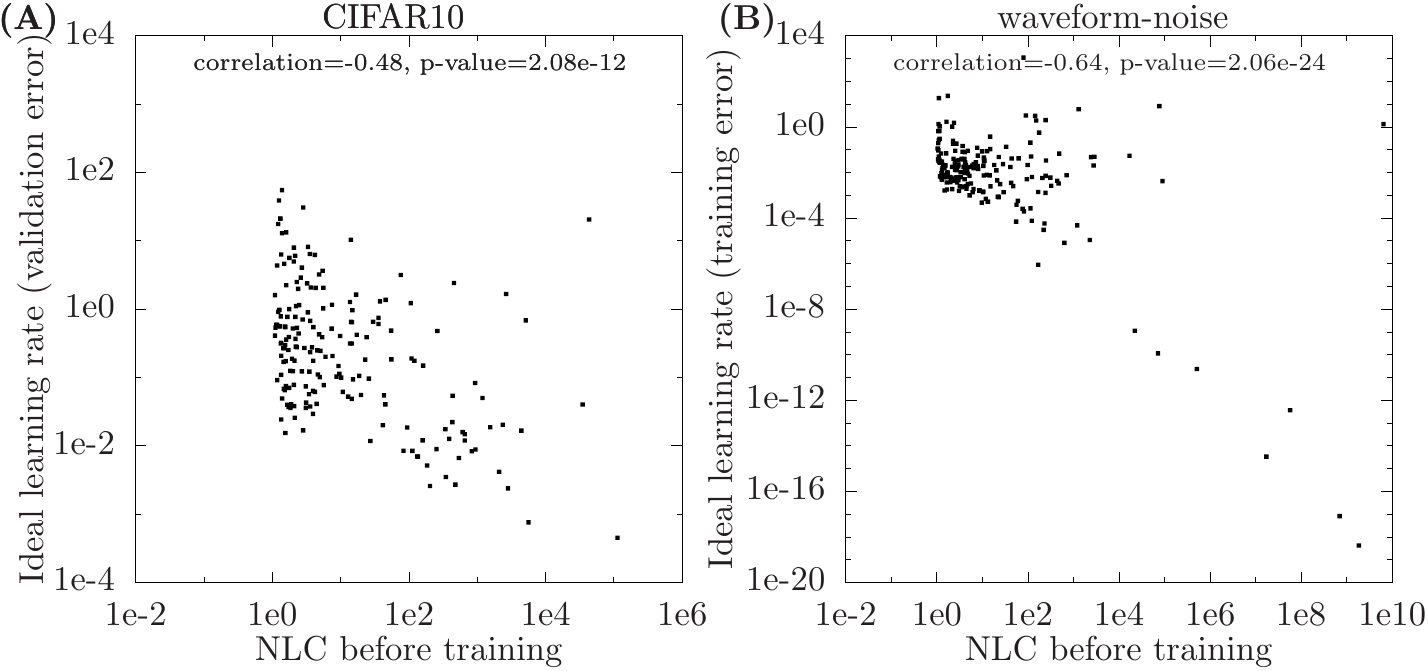}
\caption{Starting learning rate of the selected training run for minimizing validation error on CIFAR10 (A) and minimizing training error on waveform-noise (B). Note: Architectures which did not achieve a better-than-random test error were omitted in (A) and architectures that did not achieve a better-than-random training error were omitted in (B). We set those thresholds at 80\% for CIFAR10 (10 different labels) and 50\% for waveform-noise (3 different labels).}\label{lrate}
\end{figure*}

One of the hallmarks of our study was the fact that we conducted an exhaustive search over the starting learning rate for training with SGD. We trained our 750 architectures with 40 different starting learning rates each. Those learning rates formed a geometric sequence with spacing factor 3. The sequence was not the same for each architecture. In fact, the smallest of the 40 learning rates was chosen so that the weight update could still be meaningfully applied in 32 bit precision. See section \ref{nlcTestDetails} for details. Of course, this was simply a heuristic, with the aim of providing a range of learning rates that would contain the ideal learning rate with very high probability. 

To verify that this goal was achieved, in figure \ref{trainrun}A, we plot a histogram of the index of the training run that yielded the lowest validation error for CIFAR10. The training run with index 1 used the lowest starting learning rate, whereas the training run with index 40 used the largest starting learning rate. Note that we did not plot architectures that did not attain a test error of under 80\%, i.e. a non-random test error, as for those architectures the learning rate was not chosen meaningfully. We find that while a wide range of training run indeces were chosen, there was a wide margin on each side of training runs that were never chosen. This is precisely what confirms that, with high probability, we found the ideal learning rate for each architecture that has the potential to generalize.

We also retrained our waveform-noise architectures without applying early stopping based on the validation error. Instead, we continued training to determine the lowest training classification error that could be achieved. The results were plotted in figure \ref{NLCcombo}D and \ref{bias}D. For this experiment, we used 60 training runs. Here, the smallest starting learning rate was chosen so that the weight updates could still be meaningfully applied in 64 bit precision. In figure \ref{trainrun}B, we find that indeed the range of training run indeces used is much wider. For 3 architectures, the chosen training run falls outside the range of the original 40 training runs.

We hypothesized that architectures that have very high NLCs and cannot generalize are nonetheless trainable with very small learning rates in 64 bit precision. We find this to be true at least in some cases. In figure \ref{lrate}, we plot the NLC in the randomly initialized state against the starting learning rate corresponding to the chosen training run. Figure \ref{lrate}A depicts learning rates which minimized validation error on CIFAR10 and figure \ref{lrate}B depicts learning rates which minimized training error on waveform-noise. In other words, we show the same training runs as in figure \ref{trainrun}, and again we removed architectures for which generalization / training failed completely, respectively. While the range of learning rates that lead to good generalization falls in a comparatively smaller range, some architectures can only be trained successfully with a learning rate as small as $4*10^{-19}$! 

In general, the reason for this trend is that a large NLC is associated with large gradients, and these gradients need to be down-scaled to keep weight updates bounded. Intriguingly, figure \ref{lrate}B suggests that as the NLC grows, the learning rate should decay as the {\it square} of the NLC. This observation mirrors that of \citet{expl}, who found that the magnitude of weight updates should scale inversely as the gradient increases, which would require the learning rate to scale with the inverse square.

\subsection{The importance of avoiding excessive output bias} \label{biasSection}

\begin{figure*}
\includegraphics[width=\textwidth]{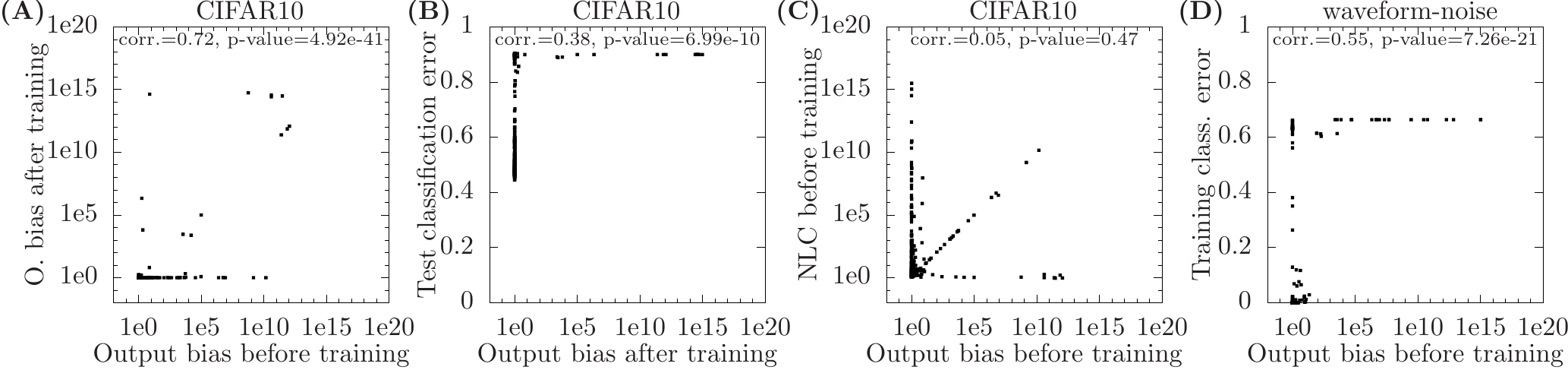}
\caption{Detailed results from our empirical study. See main text for explanation and section \ref{nlcTestDetails} for further details. All output bias and NLC values are computed on the training set.}\label{bias}
\end{figure*}

In figure \ref{NLCcombo}A, we show that high output bias before training, defined as $\sqrt{\frac{\mathbb{E}_x||f(x)||_2^2}{\mathbb{E}_x||f(x)-\bar{f}||_2^2}}$, leads to high test error. In figure \ref{bias}, we investigate this quantity further. In figure \ref{bias}A, we find that just like the NLC, the output bias decreases during training in many cases. In fact, it often reaches a value very close to 1. In figure \ref{bias}B, we find that this is in fact necessary for the network to achieve a better-than-random test error at all. This is not entirely surprising for a dataset like CIFAR10, where each label occurs equally frequently. In figure \ref{bias}C, we show that many architectures (those near the bottom of the chart) attain a high output bias but a low NLC. This confirms that a high output bias makes an independent contribution to test error prediction. Finally, in figure \ref{bias}D, we find that all architectures with a high output bias are completely untrainable on the waveform-noise dataset. This stands in contrast with figure \ref{NLCcombo}D, where we found that some high-NLC architectures were able to achieve zero or near-zero training error. Note that in figure \ref{bias}D, just as in figure \ref{NLCcombo}D, we show the lowest training error attained by retraining all architectures with 60 different starting learning rates and not using early stopping based on validation error.

Finally, we note that at the time of writing, we are working on an ``improved'' version of SGD that can successfully train high-output bias architectures and enable them to generalize. Discussing this algorithm, as well as the other signals that exist in figure \ref{bias} (e.g. many architectures cluster around 1D subspaces in graphs A/B/C), unfortunately, goes beyond the scope of this paper. 

\subsection{The value of using skip connections} \label{skipsplitSection}

\begin{figure*}[!ht]
\includegraphics[width=\textwidth]{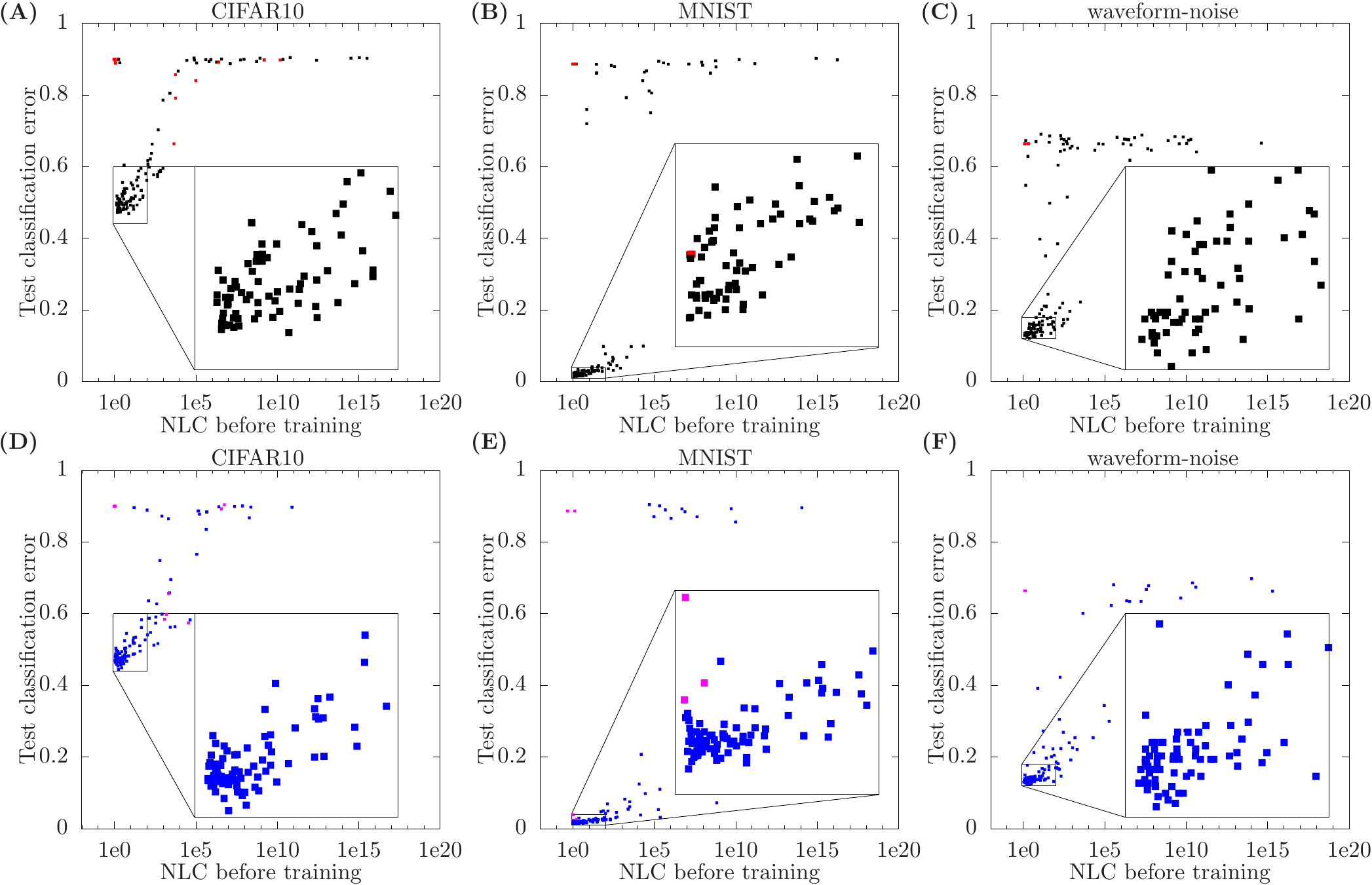}
\caption{Both the first row of graphs (A/B/C) and the second row of graphs (D/E/F) are identical to figure \ref{NLCtest}, except the top row shows only architectures without skip connections and the bottom row shows only architectures with skip connections. Again, red color indicates architectures with high output bias.}  \label{skipsplit}
\end{figure*}

In figure \ref{NLCtest}, we show in blue all architectures that have skip connections, whereas we show in black architectures without skip connections. In that figure, we find that architectures with skip connections not only exhibit a lower NLC overall, but also tend to outperform architectures without skip connections that have similar NLCs. 

As it can be hard to distinguish colors in a scatter plot, in figure \ref{skipsplit}, we plot the results for both types of architectures separately. Both the first row of graphs (A/B/C) and the second row of graphs (D/E/F) are identical to figure \ref{NLCtest}, except the top row shows only architectures without skip connections and the bottom row shows only architectures with skip connections. The differences are clear.

\section{Architecture sampling} \label{architectureSection}

In this section, we describe the randomly sampled architectures that we used for our large-scale study.

Each network layer is composed out of a fully-connected linear operation with trainable bias vector and an activation function. Some architectures have a normalization operation between the linear operation and the activation function. The last layer does not contain an activation function. Some architectures have skip connections, which always bypass two layers as in \citet{resNetTrueIdentity}. They start after either the linear operation or after the normalization operation. They end after the linear operation. The first skip connection begins after the linear or normalization operation in the first layer. The last skip connections ends after the linear operation in the last layer. All skip connections are identity skip connections, except the last skip connection, which has different input and output widths $d_\text{skip\_in}$ and $d_\text{skip\_out}$ respectively. The last skip connection multiplies the incoming signal with a $d_\text{skip\_out} \times d_\text{skip\_in}$ submatrix of a $\max(d_\text{skip\_in},d_\text{skip\_out}) \times \max(d_\text{skip\_in},d_\text{skip\_out})$ uniformly random orthogonal matrix, multiplied by $\max(1,\sqrt{\frac{d_\text{skip\_out}}{d_\text{skip\_in}}})$. The multiplier is chosen to approximately preserve the scale of the incoming signal in the forward pass. This projection matrix is not trained and remains fixed throughout training.

Each architecture was selected independently at random via the following procedure.

\begin{itemize}
\item depth: Depth is chosen uniformly from the set of odd numbers between and including 3 and 49. We used odd numbers to avoid conflicts with our skip connections, each of which bypass two linear operations but do not bypass the first linear operation.
\item width: Width was chosen automatically as a function of depth so that the number of trainable parameters in the network is approximately 1 million. The width of all layers except the input and output layer, which are determined by the data, is identical.
\item linear operation: A $d_\text{outgoing} \times d_\text{incoming}$-dimensional weight matrix is initialized as a $d_\text{outgoing} \times d_\text{incoming}$-submatrix of a $\max(d_\text{incoming},d_\text{outgoing}) \times \max(d_\text{incoming},d_\text{outgoing})$ uniformly random orthogonal matrix, multiplied by $\max(1,\sqrt{\frac{d_\text{outgoing}}{d_\text{incoming}}})$. The advantages of orthogonal over Gaussian matrices have been documented by e.g. \citet{orthogonalInitialization,eigenspectrumGram,orthRNN,fastfoodUnitaryRNN,meanFieldCNN,eigenspectrum}. We used the multiplier of $\max(1,\sqrt{\frac{d_\text{outgoing}}{d_\text{incoming}}})$ so that the scale of the signal is approximately preserved as it passes forward through the weight matrix, which is a well-accepted practice for avoiding exponential growth or decay in the forward pass used in e.g. He initialization \citep{heInit} and SELU initialization \citep{selu}. With a probability of 50\%, we initialize all trainable bias vectors as zero vectors and with a probability of 50\%, we initialize their components as independent zero mean Gaussians with a variance of 0.05. We took the 0.05 value from \citet{depthScalesMeanField}. If the bias vectors are initialized as nonzero, we scale the weight matrices with a factor of $\sqrt{0.95}$ to approximately preserve the scale of the output of the entire linear operation. Finally, with a 25\% probability, we then additionally multiply all weight matrices and bias vectors jointly by 0.9 and with a 25\% probability, we multiply them by 1.1.
\item normalization: With a 50\% probability, no normalization is used. With a 25\% probability, batch normalization \citep{batchNormalization} is used. With a 25\% probability, layer normalization \citep{layerNormalization} is used. Normalization operations do not use trainable bias and variance parameters.
\item activation function: We select one of the 8 activation functions shown in figure \ref{nlinfo}. We select ReLU, SELU and Gaussian with probability $\frac{2}{11}$ each and tanh, even tanh, sigmoid, square and odd square with probability $\frac{1}{11}$ each. We downweighted the probabilities of tanh, even tanh and sigmoid as we considered them similar. The same holds for square and odd square. After choosing the initial activation function, we added additional modifications. If the initial activation function is $\tau(s)$, we replace it by $c(\tau(ds+t)+b)$. First, $d$ and $t$ are chosen. $d$ is 1 with a 50\% probability, 1.2 with a 25\% probability and 0.8 with a 25\% probability. $t$ is 0 with a 50\% probability, 0.2 with a 25\% probability and -0.2 with a 25\% probability. Then, with a 50\% probability, we set $b$ to 0 and with a 50\% probability, we set $b$ so that if $s$ follows a unit Gaussian distribution, $\tau(ds+t)+b$ is unbiased, i.e. $\mathbb{E}_{s\sim \mathcal{N}(0,1)}\tau(ds+t)+b = 0$. Debiasing follows the example of \citet{normProp}. Finally, we always set $c$ so that if $s$ is a unit Gaussian, then $\mathbb{E}_s(c(\tau(ds+t)+b))^2 = 1$. Again, this follows the principle of avoiding exponential growth / decay in the forward pass as mentioned above. $d$, $b$, $c$ and $t$ are fixed throughout training.
\item skip connections: With a 50\% probability, no skip connections are used. With a 25\% probability, skip connections of strength 1 are used, as is usually done in practice. With a 25\% chance, we choose a single value uniformly at random between 0 and 1 and set the strength of all skip connections to that value. With a 50\% chance, all skip connections start after the linear operation. With a 50\% chance, they start after the normalization operation. We introduced these variations to obtain a more diverse range of NLCs amongst networks with skip connections. Note that normalizing the signal between skip connections rather than only within a skip block reduces the gradient damping of the skip connections for reasons related to the $k$-dilution principle \citep{expl}.
\end{itemize}

After sampling, we apply one step of post-processing. All networks that have square or odd square activation functions, or skip connections, that also do not have normalization were assigned either batch normalization or layer normalization with 50\% probability each. This is, again, to avoid exponential instability in the forward pass. This post-processing lead to the following changes in aggregate frequencies: no normalization - 20.4\%, batchnorm - 39.8\%, layer norm - 39.8\%.

We sampled 250 architectures for each of three datasets. Results pertaining to those architectures are shown in figures \ref{sensi}, \ref{NLCtest}, \ref{NLCcombo}, \ref{depth}, \ref{trainrun}, \ref{lrate}, \ref{bias} and \ref{skipsplit}.

We used softmax+cross-entropy as the loss function, as is done in the overwhelming number of practical cases. Crucially, after initializing each architecture, we measured the scale $c$ of activations fed into the loss function, i.e. $c = \sqrt{\frac{\mathbb{E}_{x} ||f(x)||_2^2}{d_\text{out}}}$. We then had the loss function divide the incoming activations by $c$ before applying softmax. This was done so that the loss functions, which yields very different training dynamics when presented with inputs of different sizes, did not confound the outcomes of our study. We believe that the preference of softmax+cross-entropy for outputs of a certain size has confounded the results of studies in the past. $c$ remained fixed throughout training.

When designing our sampling scheme, we attempted to strike a balance between relevance and diversity. On the one hand, we did not want to include architectures that are pathological for known reasons. We initialized all architectures so that the signal could not grow or decay too quickly in the forward pass. Also, we always used orthogonal initialization. The advantages of orthogonal initialization over Gaussian initialization, at least for fully-connected layers has, in our opinion, been demonstrated to the point where we believe this should be the default going forward. 

On the other hand, we introduced many variations such as activation function dilation and shift, and skip connection strength that made our architectures more diverse. While those variations are not necessarily common in practice, we made sure that we never deviated from the ``default case'' by a large amount in any particular area.

\section{Datasets} \label{datasetsSection}

\subsection{Selection}

We wanted to conduct experiments on three different datasets. First, we chose MNIST and CIFAR10 as they are the two most popular datasets for evaluating deep neural networks, and are small enough so that we could conduct a very large number of training runs with the computational resources we had available. The MNIST dataset is composed of 28 by 28 black and white images of handwritten digits associated with a digit label that is between 0 and 9 (citation: \citet{MNIST}). The CIFAR10 dataset is composed of 32 by 32 color images of objects from 10 categories associated with a category label (citation: \citet{CIFAR10}).

We decided to choose our third dataset from the UCI repository of machine learning datasets. \citet{selu} recently validated the SELU activation function, which has since become somewhat popular, on a large number of datasets from this repository. We wanted to choose a dataset that \citet{selu} also used. To decide upon the specific dataset, we applied the following filters:

\begin{itemize}
\item The most frequent class should not be more than 50\% more frequent than the average class.
\item The dataset should contain between 1.000 and 100.000 datapoints.
\item Datapoints should contain at least 10 features.
\item The dataset should not be composed of images, as we already study 2 image datasets.
\item The dataset should not contain categorical or very sparse features.
\item We only considered datasets that we were actually able to find on the repository website.
\end{itemize}

After applying all these filters, we were left with two datasets: waveform and waveform-noise. They are very similar. We chose the latter because of the greater number of input features. The inputs of the waveform-noise dataset are composed of wave attributes. Each input is associated with one of three category labels based on the wave type (citation: \citet{waveform-noise}).

\subsection{Processing}

For waveform-noise, we normalized the mean and variance of the features. We processed CIFAR10 via the following procedure.

\begin{enumerate}
\item We normalize the mean and variance of each data input. 
\item We normalize the mean of each feature.
\item Via PCA, we determine the number of dimensions that hold 99\% of the variance. That number is 810. 
\item We map each data input to an 810-dimensional vector via multiplication with a $3072 \times 810$ submatrix of a $3072 \times 3072$ uniformly random orthogonal matrix. 
\item Finally, we multiply the entire dataset with a single constant so that we obtain $\frac{\mathbb{E} ||x||_2^2}{d_\text{in}} = 1$.
\end{enumerate}

We used the exact same procedure for MNIST, except that the number of dimensions of the final dataset was 334 instead of 810.

During preliminary experiments, we found that this pre-processing scheme lead to faster training and lower error values than training on the raw data where only the features are normalized. The reason we designed this scheme in the first place was to reduce input dimensionality so that we could avoid an excessive amount of computation being allocated to the first layer, which would strain our computational budget.

The MNIST dataset contains 60.000 training data points and 10.000 test data points. The training data was randomly split into a training set of size 50.000 and validation set of size 10.000. The CIFAR10 dataset contains 50.000 training data points and 10.000 test data points. The training data was randomly split into a training set of size 40.000 and a validation set of size 10.000. The waveform-noise dataset contains 5.000 data points and was randomly split into a training set of size 3.000, a validation set of size 1.000 and a test set of size 1.000.

As mentioned, for CIFAR10, our input dimensionality was 810. For MNIST, it was 334. For waveform-noise, it was 40. For CIFAR10 and MNIST, the output dimensionality / number of classes was 10. For waveform-noise, it was 3.

\section{Experimental details} \label{detailsSection}

\subsection{NLC vs nonlinearity study (figure \ref{sensi})} \label{sensiDetails}

We compute the NLC as defined in section \ref{derivationSection} using the algorithm from section \ref{computingSection}.

We compute the median of the nonlinearity distribution as follows. First, note that we alter the definition from section \ref{derivationSection} somewhat by considering the local linear approximation not at an individual input $x$ but at an entire batch. This was done to facilitate architectures with batch normalization where $f$ is only defined over batches. We also set the tolerance $T = 2$. 

We sample a single value $C$ from the nonlinearity distribution as follows. We sample a data batch $X \in \mathbb{R}^{d_\text{in} \times B}$ of batch size $B$. Each column of $X$ contains a random data input from the training set, sampled without replacement. We also draw an input direction matrix $U \in \mathbb{R}^{d_\text{in} \times B}$ and output direction matrix $V \in \mathbb{R}^{d_\text{out} \times B}$ where each column of $U$ is independently drawn from $\mathcal{N}(0,\Cov_x)$ and each column of $V$ is independently drawn from $\mathcal{N}(0,I_{d_\text{out}})$. $\Cov_x$ is pre-computed once for each of our three datasets. We initially set $c = 10^{-9}$ and then check whether the condition $\frac{c}{2}\sum_{1\le i\le d_\text{in},1\le j\le d_\text{out}, 1\le k,l\le B}V_{jk}\mathcal{J}(X)_{jkil}U_{il} \le \sum_{1\le j\le d_\text{out}, 1\le k\le B}(f(X + cU) - f(X))_{jk}V_{jk} \le 2c\sum_{1\le i\le d_\text{in},1\le j\le d_\text{out}, 1\le k,l\le B}V_{jk}\mathcal{J}(X)_{jkil}U_{il}$ holds for increasing values of $c$ until the condition fails. Here $f$ and $\mathcal{J}$ can be taken to be applied independently to each column of $X$ if the network does not use batchnorm and taken to be applied jointly to all inputs in $X$ if the network does contain batchnorm. $\mathcal{J}(X)_{jkil}$ denotes the 4D tensor that describes the derivative of each component of $f(X)$ with respect to each component of $X$. The tensor product $V\mathcal{J}(X)U$ can be evaluated cheaply by forward-propagating $X$, clamping $V$ to the output layer, backpropagating $V$ to the input layer, and finally computing the inner product between the result of this computation and $U$. The values of $c$ we checked formed a geometric series with spacing factor $10^{\frac{1}{10}}$, which is around $1.26$. We could not reliably check values of $c$ less than $10^{-9}$ due to numerical underflow, which is why architectures with an NLC less than $10^{-9}$ are not shown in figure \ref{sensi}.

We use a total of 10 batches of size 250 from the respective dataset and draw 10 random $U$ and 10 random $V$. We obtain one sample of $C$ for each of $10*10*10 = 1000$ configurations. Finally, in figure \ref{sensi}, we report the median across those 1000 values for each architecture.

For each architecture, we considered a single random initialization for computing both the median nonlinearity and the NLC. All values were computed in the randomly initialized state. No training was conducted.

\subsection{Predictiveness study (figures \ref{NLCtest}, \ref{NLCcombo}, \ref{depth}, \ref{trainrun}, \ref{lrate}, \ref{bias} and \ref{skipsplit})} \label{nlcTestDetails}

For each architecture, we considered a single random initialization. We trained each architecture with SGD using batches of size 250. To ensure that there is no bias with regards to learning rate, we tuned the starting learning rate independently for each architecture by conducting a large number of training runs with various starting learning rates. A training run is conducted as follows. We train with the starting learning rate until the validation classification error (VCE) has not decreased for 10 epochs. Then we rewind the state of the network by 10 epochs (when the lowest VCE was achieved), divide the learning rate by 3 and continue training until the VCE has not improved for 5 epochs. We divide the learning rate by 3 again, rewind and continue training until the VCE has not improved for 5 epochs again. This process continues until the learning rate has been divided by 3 ten times. When the VCE has again not improved for 5 epochs, we rewind one last time and terminate the training run.

For each architecture we completed 40 total training runs with 40 different starting learning rates that form a geometric series with spacing factor 3. For each architecture, the smallest starting learning rate considered was computed as follows. We ran the SGD optimizer for 1 epoch with a learning rate of 1 without actually applying the updates computed. For the weight matrix in each layer, we thus obtained one update per batch. Let $\delta W_{lb}$ denote the update obtained for layer $l$ and batch $b$ and let $W_l$ denote the initial value of the weight matrix in layer $l$. Finally, we used the value $10^{-8}\sum_l\frac{\sqrt{\mathbb{E}_{b}||\delta W_{lb}||^2_F}}{||W_l||_F}$ as our smallest starting learning rate. The rational behind this choice was that no individual weight matrix update obtained with the smallest starting learning rate would perturb any weight matrix during any iteration by more than approximately $10^{-8}$. We chose $10^{-8}$ specifically so that our smallest starting learning rate would be less than the smallest learning rate that can be meaningfully used under 32 bit precision. Nonetheless, we trained all networks using 64 bit precision. 

Of course, this choice of smallest starting learning rate is merely a heuristic. The goal of this heuristic is to ensure that the best starting learning rate for each architecture is within the range of starting learning rates considered with very high probability. We validated this heuristic by checking that no architecture that obtained a non-random test classification error attained its lowest VCE with either the smallest five or largest five starting learning rates considered. This condition was fulfilled for all architectures and datasets. See section \ref{lratesection} for details. Henceforth, we refer to the `trained network' as the network that was obtained after the training run that yielded the lowest VCE and the `initial network' as the network in the randomly initialized state.

In figure \ref{trainrun}A, we show which training runs achieved the lowest VCE and in figure \ref{lrate}A, we show which starting learning rates achieved the lowest VCE, plotted against the NLC of the initial network.

In figure \ref{NLCtest}, we plot the NLC of the initial network against the test error of the trained network. We mark in red all points corresponding to architectures for which $\sqrt{\frac{\mathbb{E}_x||f||_2^2}{\mathbb{E}_x||f-\bar{f}||_2^2}} > 1000$ for the initial network. We mark in blue all points corresponding to architectures that have skip connections. In figure \ref{depth}, we plot depth versus test error of the trained network.

In figure \ref{NLCcombo}A, we plot the output bias value $\sqrt{\frac{\mathbb{E}_x||f(x)||_2^2}{\mathbb{E}_x||f(x)-\bar{f}||_2^2}}$ of the initial network against the test error of the trained network. In figure \ref{NLCcombo}B, we plot the NLC of the initial network against the NLC of the trained network. If figure \ref{NLCcombo}C, we plot the NLC of the trained network against the test error of the trained network. In both \ref{NLCcombo}B and \ref{NLCcombo}C, the NLC was computed on the training set. However, the value of the NLC computed on the test set was very similar. We further compare the output bias of the initial network against the output bias of the trained network, against test error and against the NLC of the initial network in figure \ref{bias}. Finally, in figure \ref{skipsplit}, we break down the results of figure \ref{NLCtest} into architectures with skip connections and architectures without skip connections.

The NLC and output bias were computed on the training set according to the algorithms given in sections \ref{computingSection} and \ref{biasComputingSection} respectively.

We then re-trained our waveform-noise architectures with two changes to the protocol: We reduced the learning rate by a factor of 3 only once the training classification error (TCE) had not been reduced for 10 / 5 epochs respectively; and we considered 60 different learning rates which formed a geometric series with spacing factor 3 and start value $10^{-16}\sum_l\frac{\sqrt{\mathbb{E}_{b}||\delta W_{lb}||_F^2}}{||W_l||_F}$. Therefore, we considered even the smallest starting learning rate that was meaningful for 64 bit precision training. This change allowed us to successfully train even architectures with very high NLCs. See section \ref{lratesection} for an analysis on this point. The reason we only re-trained the waveform-noise architectures for this scenario is because training can take a very long time without using the validation set for early stopping, leading to considerable computational expense.

Again, the goal of our heuristic is to ensure that the best starting learning rate for each architecture was within the range considered with very high probability. We verified this again by checking that no architecture which obtained a non-random TCE attained its lowest TCE with either the smallest or largest five starting learning rates. This condition was fulfilled for all architectures. Note that if we had used the original set of 40 starting learning rates, this check would have failed. See section \ref{lratesection} for details.

The lowest TCE achieved is plotted against the NLC and output bias of the initial network in figures \ref{NLCcombo}D and \ref{bias}D respectively. In figure \ref{trainrun}B we show which training runs yielded the lowest TCE and in figure \ref{lrate}B we show which starting learning rates yielded the lowest TCE, plotted against the NLC of the initial network.

Finally, for figure \ref{NLCcombo}F, we re-trained our 250 waveform-noise architectures with Adam instead of SGD. The protocol was the same (40 training runs), except before obtaining our measurements for $\delta W_{lb}$, we first ran Adam for 4 epochs, again without applying updates, in order to warm-start the running averages. Only then did we run it for another epoch where we actually gathered values for $\delta W_{lb}$. Again, we verified that the first and last 5 training runs were never used.

\subsection{Error robustness study (figure \ref{NLCcombo})} \label{errorRobustnessDetails}

We computed the maximal error-preserving perturbation shown in figure \ref{NLCcombo}E similarly to the median nonlinearity in section \ref{sensiDetails}. The difference is that instead of requiring that the local linear approximation be close to the true function, we required that the test error over the path from $X$ to $X + cU$ be at most 5\% higher than than the test error at $X$. The test error ``over the path'' is defined as the fraction of inputs in the batch that were incorrectly classified somewhere on the line segment from $X$ to $X + cU$. Again, we started with $c=10^{-9}$ and increased it by $10^{\frac{1}{10}}$ at each step, checking whether each input is correctly or incorrectly classified. We chose the 5\% threshold so that architectures with a test error of around 90\% on CIFAR10 / MNIST would yield finite outcomes. The values shown in figure \ref{NLCcombo}E are the median over $10*10=100$ values obtained from 10 random batches of size 250 and 10 Gaussian random direction matrices $U$. The random direction matrix $V$ used in section \ref{sensiDetails} does not come into play here.

\subsection{Approximability study (table \ref{nlinfo})} \label{approximabilityDetails}

$NLC_\tau$ was computed as defined in section \ref{meaningSection}. $NLC_\tau^{48}$ is simply the exponentiated value. The linear approximation error is computed as $\frac{\mathbb{E}_{s \sim \mathcal{N}(0,1)}(\tau(s)-\bar{\tau}(s))^2}{\mathbb{E}_{s \sim \mathcal{N}(0,1)}\bar{\tau}(s)^2}$, where $\bar{\tau}$ is the best linear fit to $\tau$ for inputs drawn from $\mathcal{N}(0,1)$, i.e. $\arg \min_{\bar{\tau} \text{ linear}} \mathbb{E}_{s \sim \mathcal{N}(0,1)}(\tau(s)-\bar{\tau}(s))^2$. 

NLC was computed as in section \ref{computingSection}. We show the median across 10 random initializations. The values for different initializations show little variation except for 49-layer networks with square or odd square activation functions.

\subsection{Confounder study (figure \ref{conf})} \label{confDetails}

We began by training a 5-layer batchnorm-ReLU network. The layout of the network in terms of the sequence of linear operations, batchnorm operations and ReLU operations was as in section \ref{architectureSection}. The width of the input layer was 40 and the width of the output layer was 3 according to the dimensionality of data inputs / number of classes in the waveform-noise dataset. The width of intermediate layers was 100. Each weight matrix was initialized as a sub-matrix of an orthogonal matrix as described in section \ref{architectureSection}, except that these matrices were additionally scaled by a factor of $\sqrt{2}$. This is the popular He initialization \citep{heInit}. 

We trained this network with SGD using the protocol outlined in section \ref{nlcTestDetails} involving 40 starting learning rates and learning rate decay. The ideal starting learning rate selected was 57.9 and the corresponding test error was 13\%. 

We then retrained the same network in four different ways. (A) We multiplied the input data with various constants $c$. For this scenario, we used starting learning rate 57.9. (B) We multiplied the loss function with various constants $c$ and then used starting learning rate $\frac{57.9}{c}$. (C) We replicated each input dimension $c$ times for various values of $c$. In this scenario, because the input dimensionality increased, we correspondingly reduced the initial scale of each component in the first weight matrix by $\sqrt{c}$ as dictated by He initialization. In that scenario, we also used a starting learning rate of 57.9 for all weight matrices except the first weight matrix, where we used $\frac{57.9}{c}$. We scaled the learning rate of the first weight matrix to maintain the overall amount of training the first layer receives as the input dimensionality grows. (D) We added various constants $c$ to the input data and set the starting learning rate applied to the first weight matrix to zero while using a starting learning rate of 57.9 for all other weight matrices.

We present results in figure \ref{conf}A/B/C/D respectively. In scenarios A, B and D, the test error was completely unchanged as $c$ varied. In fact, the entire learning trajectory remained the same. Note that in scenario D, because the first layer does not learn, the overall error is slightly elevated. In scenario C, the test error remained almost unchanged.

We then built analogous plain ReLU He-initialized networks (without batchnorm) of different depths and trained then with our 40-learning rate protocol. Results are shown in figure \ref{conf}E. Finally, we built analogous plain 2-layer networks with the sawtooth nonlinearity, which is illustrated in figure \ref{saw}. It is defined as $\tau(s) = p(\frac{s}{p} - \lfloor\frac{s}{p}\rfloor)$ when $\frac{s}{p} - \lfloor\frac{s}{p}\rfloor < 0.25$, as $\tau(s) = p(\frac{s}{p} - \lfloor\frac{s}{p}\rfloor - 1)$ when $\frac{s}{p} - \lfloor\frac{s}{p}\rfloor > 0.75$ and $\tau(s) = p(0.5 - \frac{s}{p} + \lfloor\frac{s}{p}\rfloor)$ otherwise. We used different sawtooth periods $p$. The weight matrices were initialized as in section \ref{architectureSection}, which makes sense given that $\mathbb{E}_{s \sim \mathcal{N}(0,1)}\tau_\text{sawtooth}(s)^2 \approx 1$ for all periods.

For every experiment, we used a single random initialization. The dataset used was waveform-noise.

We computed the following metrics, all in the randomly initialized state with expectations taken over the training set.

\begin{itemize}
\item NLC: Defined in section \ref{derivationSection} and computed as in section \ref{computingSection}.
\item Gradient vector component size: $\sqrt{\mathbb{E}_x\frac{||\frac{d\ell}{dx}||_2^2}{d_\text{in}}}$
\item Gradient vector length: $\sqrt{\mathbb{E}_x||\frac{d\ell}{dx}||_2^2}$
\item Change in Lipschitz constant: The Lipschitz constant is defined as $\max_{x,x' \in \mathbb{D}} \frac{||f(x)-f(x')||_2}{||x-x'||_2}$, where $\mathbb{D}$ is an unspecified input domain. We assume that as the data inputs are scaled with a constant $c$, so is $\mathbb{D}$. Values shown are relative to the value obtained for $c=1$, which we do not compute.
\item Input correlation: \\$\sqrt{\mathbb{E}_{x,x'}\frac{((x-\bar{x})^T(x'-\bar{x}))^2}{((x-\bar{x})^T(x-\bar{x}))((x'-\bar{x})^T(x'-\bar{x}))}}$
\item Output correlation: \\$\sqrt{\mathbb{E}_{x,x'}\frac{((f(x)-\bar{f})^T(f(x')-\bar{f}))^2}{((f(x)-\bar{f})^T(f(x)-\bar{f}))((f(x')-\bar{f})^T(f(x')-\bar{f}))}}$
\end{itemize}

\section{Computing the NLC} \label{computingSection}

The NLC can be cheaply, stochastically computed. First, we examine the case where the network $f$ does not contain batch normalization, and is thus defined independently for each input $x$. Specifically, we begin by describing how to compute the denominator of the NLC, $\Tr(\Cov_f)$. We notice that $\Tr(\Cov_f) = \mathbb{E}_{x\sim \mathcal{D}} ||f(x) - \bar{f}||_2^2$. Hence, we first compute $\bar{x}$ and $\bar{f}$ exactly via a single pass over the dataset. $\bar{x}$ is simply the mean of all data inputs and $\bar{f}$ is the mean of the outputs corresponding to those data inputs. We then compute $\Tr(\Cov_f) = \mathbb{E}_x ||f(x) - \bar{f}||_2^2$ exactly via a second pass over the dataset. For each data input, we first compute $f(x)$ via forward propagation and then immediately subtract the pre-computed value $\bar{f}$ from that value before computing the squared length. 

Alternatively, it is possible to use only a single pass during which $ \mathbb{E}_x ||f(x)||_2^2$ and $\bar{f}$ are computed and then to set $\Tr(\Cov_f)$ as $[\mathbb{E}_x ||f(x)||_2^2] - ||\bar{f}||_2^2$. However, this method can only compute $\Tr(\Cov_f)$ accurately when we have $\sqrt{\frac{\mathbb{E}_x||f(x)||_2^2}{\mathbb{E}_x||f(x)-\bar{f}||_2^2}} \gtrapprox 2^{\frac{b}{2}}$, where $b$ is the number of bits used in the floating point computation. Using two passes over the dataset allows one to compute $\Tr(\Cov_f)$ when $\sqrt{\frac{\mathbb{E}_x||f(x)||_2^2}{\mathbb{E}_x||f(x)-\bar{f}||_2^2}} \gtrapprox 2^b$. In plain words, if the output bias of the network is large, the one-pass method often suffers from numerical underflow when the two-pass method does not. When the computation is performed with 32 bit or even 16 bit precision, this is important even for moderate levels of output bias. For our study, it is important to compute the NLC accurately for architectures with an output bias up to $10^{16}$, which is only possible with the two-pass method and double precision. (See figures \ref{NLCcombo} and \ref{bias} for output bias values.)

Now let's look at computing the numerator of the NLC. We notice that $\mathbb{E}_{x \sim \mathcal{D}}\Tr(\mathcal{J}(x)\Cov_x\mathcal{J}(x)^T) = \mathbb{E}_{x,x',u \sim \mathcal{N}(0,I_{d_\text{out}})} (u^T[\mathcal{J}(x)](x' - \bar{x}))^2$. We already computed $\bar{x}$ previously, so we can treat this value as a fixed constant. We will compute the value of the numerator stochastically by sampling random triplets $(u,x,x')$ and then averaging over the resultant values of $(u^T[\mathcal{J}(x)](x' - \bar{x}))^2$. We first sample a random $x$ and propagate it forward through the network. We then clamp a Gaussian random vector $u$ to the output layer and backpropagate it to the input layer. The result of this computation yields exactly $u^T\mathcal{J}(x)$. Finally, we sample a second input value $x'$ and take the squared inner product of $x' - \bar{x}$ and $u^T\mathcal{J}(x)$. In practice, we parallelize this computation by using batches of size 250 and clamping Gaussian random matrices with 250 columns to the output layer. We compute the stochastic average over all 250 inputs in a batch and 100 different, independently drawn batches. Each data input in each batch is associated with an independently drawn $u$ and $x'$. Note that while our computation of $\bar{x}$, $\bar{f}$ and $\Tr(\Cov_f)$ involves a full pass over the dataset, if the dataset is very large, those quantities can of course also be computed stochastically. 

Now we examine the case of $f$ containing batch normalization. Then, the original definition of the NLC does not technically apply. We generalize the definition as:

\begin{equation*}
NLC(f,\mathcal{D},B) = \sqrt{\frac{\mathbb{E}_{X \sim \mathcal{D}^B}\Tr(\mathcal{J}(X)\Cov_x^B\mathcal{J}^T(X))}{\Tr(\Cov_f^B)}}
\end{equation*}

Here, $X$ is the data batch represented by a $d_\text{in} \times B$ matrix where each column corresponds to a random data input, drawn without replacement. $\Cov_x^B$ is the $Bd_\text{in} \times Bd_\text{in}$ covariance matrix of $X$. $\Cov_f^B$ is the $Bd_\text{in} \times Bd_\text{in}$ covariance matrix of $f(X)$, which is the $d_\text{out} \times B$ matrix where columns correspond to the outputs obtained when the data inputs corresponding to the columns of $X$ are propagated jointly through the network. $\mathcal{J}(X)$ is a $Bd_\text{out} \times Bd_\text{in}$ matrix representing the derivative of all components of $f(X)$ with respect to all components of $X$. Note that the NLC now depends on the batch size $B$. We set $B=250$ as this is the value we used for network training. It is easy to check that this value defaults to the original definition of the NLC when batch normalization is not used.

In practice, we used the exact same procedure for computing the NLC with batch normalization as we did for computing the NLC without batch normalization. We still have $\Tr(\Cov_f^B) = \mathbb{E}_X ||f(X)-\bar{f}||_2^2$, where $\bar{f}$ is the mean of all outputs encountered using batches of size 250 and is subtracted column-wise from $f(X)$. The estimate obtained by computing $\bar{f}$ and then $\Tr(\Cov_f^B)$ with two passes over the dataset is a low-variance version of the vanilla stochastic estimator. Similarly, we still have $\mathbb{E}_{x \sim \mathcal{D}}\Tr(\mathcal{J}(X)\Cov_X^B\mathcal{J}(X)^T) = \mathbb{E}_{X,X',U \sim \mathcal{N}(0,I_{Bd_\text{out}})} (U^T[\mathcal{J}(X)](X' - \bar{x})_{Bd_\text{in}})^2$. Here, $\bar{x}$ is subtracted column-wise from $X'$ and the $Bd_\text{in}$ subscript indicates a reshaping of the matrix into a column vector of this length. We obtain the vanilla stochastic estimator by forward-propagating batches $X$, backpropagating Gaussian random matrices $U$ clamped to the output layer, and computing the squared inner product of the result of that computation with $X' - \bar{x}$.

Finally, note that using a non-exact method for computing $\Tr(\Cov_f^B)$ will, again, lead to problems if the output bias is high. However, in practice, using batch normalization ensures that the output bias is 1 in the randomly initialized state and remains sufficiently small throughout training.

\section{Computing output bias} \label{biasComputingSection}

While computing the quantity $\sqrt{\frac{\mathbb{E}_{x\sim\mathcal{D}}||f(x)||_2^2}{\mathbb{E}_{x\sim\mathcal{D}}||f(x)-\bar{f}||_2^2}}$ is essentially trivial, as in section \ref{computingSection}, we point out that it can be necessary to first compute $\bar{f}$ exactly via a first pass over the dataset and then to compute $\mathbb{E}_x||f(x)-\bar{f}||_2^2$ exactly in a second pass by subtracting $\bar{f}$ from $f(x)$ before the squared length is computed. This allows one to compute output biases when $\sqrt{\frac{\mathbb{E}_x||f(x)||_2^2}{\mathbb{E}_x||f(x)-\bar{f}||_2^2}} \gtrapprox 2^b$, where $b$ is the number of bits used in the floating point computation. Conversely, if $\mathbb{E}_x||f(x)-\bar{f}||_2^2$ is computed as $\mathbb{E}_x||f(x)||_2^2-||\bar{f}||_2^2$ where both quantities are computed in a single pass over the dataset, we experience numerical underflow when $\sqrt{\frac{\mathbb{E}_x||f(x)||_2^2}{\mathbb{E}_x||f(x)-\bar{f}||_2^2}} \gtrapprox 2^{\frac{b}{2}}$. When the computation is performed with 32 bit or even 16 bit precision, this is important even for moderate levels of output bias. We use batches of size 250.

As in section \ref{computingSection}, if $f$ uses batch normalization, we generalize the output bias as $\sqrt{\frac{\mathbb{E}_{X \sim \mathcal{D}^B}||f(X)||_2^2}{\mathbb{E}_{X \sim \mathcal{D}^B}||f(X)-\bar{f}||_2^2}}$. Here, $X$ is the data batch represented by a $d_\text{in} \times B$ matrix where each column corresponds to a random data input drawn without replacement. $f(X)$ is the $d_\text{out} \times B$ matrix where columns correspond to the outputs obtained when the data inputs corresponding to the columns of $X$ are propagated jointly through the network. Note that the output bias now depends on the batch size $B$. We set $B=250$ as this is the value we used for network training. It is easy to check that this value defaults to the original definition of the output bias when batch normalization is not used.

As in section \ref{computingSection}, we use the same computational recipe for computing output bias for networks with batch normalization as we do for networks without batch normalization, turning the exact computation into a stochastic estimator. And again this stochasticity makes it impossible to compute large output bias values accurately. This is not an issue in practice as networks with batch normalization have an output bias of 1 in the randomly initialized state and a small output bias throughout training. 

\end{document}